\documentclass[a4paper,fleqn]{cas-sc}

\usepackage[authoryear]{natbib}

\usepackage{booktabs}
\usepackage{longtable}
\usepackage{array}
\usepackage{ragged2e}
\usepackage{float}
\usepackage{tabularx}

\def\tsc#1{\csdef{#1}{\textsc{\lowercase{#1}}\xspace}}
\tsc{WGM}
\tsc{QE}
\tsc{EP}
\tsc{PMS}
\tsc{BEC}
\tsc{DE}

\makeatletter
\expandafter\def\csname __first_footerline:\endcsname{%
  \begingroup
  \small\sffamily
  {\rmfamily\itshape Preprint submitted to Elsevier}%
  \endgroup
}
\makeatother

\begin{document}
\let\WriteBookmarks\relax
\def\floatpagepagefraction{1}
\def\textpagefraction{.001}

\shorttitle{Modelling daily activity patterns via deep representation learning}

\shortauthors{ }

\title [mode = title]{Modelling daily activity patterns from mobile phone location data via deep representation learning}                      



\author[1]{Xinglei Wang}[
                        orcid=0000-0002-9824-7663
                        ]
\ead{xinglei.wang.21@ucl.ac.uk}

\affiliation[1]{organization={SpaceTimeLab, Department of Civil, Environmental and Geomatic Engineering, University College London},
                city={London},
                country={United Kingdom}}

\affiliation[2]{organization={State Key Laboratory of Information Engineering in Surveying, Mapping and Remote Sensing, Wuhan University},
                city={Wuhan},
                country={China}}

\affiliation[3]{organization={3DIMPact, Department of Civil, Environmental and Geomatic Engineering, University College London},
                city={London},
                country={United Kingdom}}

\affiliation[4]{organization={Department of Geography, University College London},
                city={London},
                country={United Kingdom}}

\affiliation[5]{organization={Department of Geography and Resource Management, The Chinese University of Hong Kong},
                city={Hong Kong SAR},
                country={China}
                }

\affiliation[6]{organization={The Alan Turing Institute},
                city={London},
                country={United Kingdom}}

\author[1]{Junyuan Liu}[orcid=0009-0009-7194-6868]
\ead{junyuan.liu.22@ucl.ac.uk}

\author[1,2]{Guangsheng Dong}[orcid=0000-0001-7676-497X]
\ead{guangshengdong@whu.edu.cn}

\author[1,3]{Zichao Zeng}[orcid=0009-0002-8975-875X]
\ead{zichao.zeng.21@ucl.ac.uk}

\author[4,5]{Stephen Law}[orcid=0000-0003-3184-572X]
\ead{stephen.law@ucl.ac.uk}

\author[1]{James Haworth}[orcid=0000-0001-9506-4266]
\ead{j.haworth@ucl.ac.uk}

\author[1,6]{Tao Cheng}[
orcid=0000-0002-5503-9813
]
\cormark[1]
\ead{tao.cheng@ucl.ac.uk}

\cortext[cor1]{Corresponding author}

\begin{abstract}
Passively collected mobile phone location data provide large-scale, longitudinal observations of human mobility but do not directly reveal activity purposes. The functional characteristics of visited locations offer useful contextual information, yet their relationship with activity purpose remains uncertain, particularly in mixed-use urban environments. We conceptualise activity pattern mining as an integrated process of representation, clustering, and interpretation, and propose the Activity Chain Encoder (ACE) for the representation stage. ACE is a self-supervised model that combines pre-trained urban embeddings with visit timing and duration and uses a Transformer to model the sequential organisation of stays. It is trained using masked activity modelling and identity-guided contrastive learning without requiring deterministic activity purpose labels. Learned daily representations are aggregated into user-level profiles, clustered, and interpreted through temporal-functional patterns and Census-derived demographic context. Applied to mobile phone app location data from London and compared with three representative methods, ACE supports the identification of six differentiated weekday activity-pattern groups characterised by distinct daily rhythms, urban functional contexts, and demographic associations. These complementary forms of evidence further support the development of empirically grounded activity-pattern personas, establishing a holistic route for deriving behaviourally meaningful population patterns from unlabelled mobile phone location data. The source code for the entire analytical pipeline developed in this study is publicly available at https://github.com/xlwang233/ACE.

\end{abstract}




\maketitle

\section{Introduction}

Understanding how people organise their daily activities is central to travel behaviour research and transportation planning \citep{axhausen1992activity,bhat1999activity}. 
Modelling daily activity patterns provides important evidence for understanding mobility needs, population heterogeneity, transport demand, and the social organisation of urban life \citep{chapin1974human}.

Travel surveys have traditionally been the main source of evidence for analysing individual activity and travel behaviour. They provide self-reported trip purposes, transport modes, socio-demographic attributes, and  behavioural motivations. However, survey-based approaches are limited by small sample sizes, short observation periods, high collection costs, and potential reporting biases \citep{stopher2007household,wang2018applying}. These limitations make it difficult to capture daily activity behaviour across the entire urban population.

Against this background, passively collected mobile phone location data have emerged as a complementary source for activity pattern analysis. Such data can provide large-scale and longitudinal observations of individual mobility in real-world settings . They have therefore been increasingly used to study travel behaviour \citep{jiang2017activity}, activity spaces \citep{chen2023sensing}, urban dynamics \citep{wang2023ups}, and population-level mobility patterns \citep{alexander2015origin}. However, the scale and continuity of mobile phone data come at the cost of behavioural meaning. Mobile phone data record where and when a device is observed but do not directly reveal the purpose of the associated trip or activity.

Inferring activity purpose from location observations is inherently uncertain. The functional characteristics of a visited place provide contextual evidence about the activities that may occur there. However, urban environments are frequently mixed-use and may simultaneously support residential, commercial, employment, retail, leisure, and transport functions \citep{yue2017measurements}. The same location may consequently be associated with several possible activities. Positioning errors in mobile phone data can further exacerbate this uncertainty by associating an observation with different nearby urban contexts \citep{wang2018applying}. The fundamental methodological challenge is therefore how to derive behaviourally meaningful population patterns from location traces when activity purposes are unobserved and the functional context of the observed locations is uncertain.

Existing studies addressing this problem can be understood through an integrated process comprising three interdependent stages: representation, clustering, and interpretation. In the representation stage, raw mobility observations or derived activity chains are transformed into numerical descriptions of individual behaviour. In the clustering stage, individuals with similar representations are grouped to identify recurrent population-level patterns. In the interpretation stage, the resulting clusters are examined through their temporal, spatial, functional, and socio-demographic characteristics. Although these operations are already present in various forms in previous activity pattern studies, their interdependence is not made explicit. What can be identified through clustering and subsequently interpreted as a behavioural pattern depends fundamentally on the information preserved during representation. We therefore conceptualise activity pattern mining from passive location data as a representation–clustering–interpretation framework and identify activity chain representation as a central methodological bottleneck within this process.

Despite considerable progress, existing approaches to representing activity patterns remain limited in three main respects. First and most fundamentally, they must approximate activity meaning when explicit purpose labels are unavailable. Some studies infer a single activity category using heuristic rules or classifiers trained on labelled travel surveys \citep{li2024multi}. Such hard assignments can obscure the mixed-use characteristics of urban environments and convey a level of certainty that the underlying observations do not support. Other approaches use TF-IDF vectors or probabilistic topic models such as Latent Dirichlet Allocation (LDA) \citep{blei2003latent} to retain multiple semantic components within a place profile \citep{shen2018profiling,cheng2018grouping,li2022understanding,sun2026latent}. These methods avoid some limitations of single-category assignment, but their representations remain structured by the selected POI vocabulary, classification system, study-area corpus, and topic specification. They therefore characterise places primarily through a predefined or locally derived semantic feature space.

Second, daily activity behaviour is inherently multidimensional. A person’s activity pattern is shaped not only by where they go, but also by when they visit, how long they stay, and what kinds of urban functions are associated with the visited places. Prior studies have shown that jointly considering spatial, temporal, and semantic dimensions can provide a more comprehensive understanding of human activity than focusing on any single dimension alone \citep{huang2016understanding,shen2018profiling,cheng2018grouping,li2022understanding}. Nevertheless, many approaches rely on handcrafted statistical summaries, semantic profiles, or spatial distributions in which temporal detail and activity duration are represented only coarsely or indirectly. Such partial representations may fail to distinguish daily routines that involve similar places but differ in their timing and duration.

Third, activity chains are sequential and context dependent. The behavioural meaning of a stay may depend on its position within the day and on the activities that precede and follow it. For example, a short stay in an eating and drinking environment may represent a different routine when it occurs between home and work than when it occurs during an evening leisure sequence. However, aggregation-based approaches commonly combine activity-level features into an unordered individual profile, thereby losing relationships among stays within the chain \citep{cheng2018grouping,li2024multi}. Recent studies have begun to adapt natural language processing techniques to activity chain modelling \citep{li2025understanding}. Yet translating activity chains into purely semantic sentences may sacrifice native geographic, temporal, or duration information.

Taken together, these limitations reveal the need for representation learning methods that can model unlabelled daily activity chains at multiple levels: by retaining informative urban context without requiring deterministic activity-purpose assignment, preserving visit timing and duration as dimensions of daily behaviour, and capturing the sequential organisation of stays. Addressing these requirements is essential for transforming passively observed location sequences into representations that support meaningful population-level pattern discovery and interpretation.

To address this gap, we propose Activity Chain Encoder (ACE), a self-supervised representation learning model for daily activity chains derived from mobile phone location data. ACE represents each stay using a pre-trained urban embedding together with temporal and duration embeddings, thereby capturing characteristics of the surrounding urban environment as well as when the stay occurs and how long it lasts. The resulting sequence is processed by a Transformer encoder to model the ordering and relationships among stays within the daily chain. ACE is trained through two complementary objectives: masked activity modelling, which reconstructs the urban embedding of masked stays from the surrounding sequence to learn contextual relationships among visited locations; and identity-guided contrastive learning, which encourages consistency among daily chains observed for the same user and day type. Together, these components produce activity-chain embeddings that integrate urban context, timing, duration, and sequential organisation without requiring ground-truth activity-purpose or activity-pattern labels.


The learned representations are operationalised through the full representation–clustering–interpretation framework. Daily activity-chain representations produced by ACE are aggregated into user-level activity profiles and clustered to identify groups with similar activity patterns.
Because the resulting clusters do not have ground-truth behavioural labels, they are interpreted through complementary evidence. Semantic probing relates the pre-trained urban embeddings to customisable urban functional categories, enabling the temporal and functional contexts associated with each cluster to be examined. Census-derived demographic characteristics associated with users’ inferred home locations provide further contextual evidence. Integrating these sources supports the development of empirically grounded activity-pattern personas that summarise the characteristic routines, visited urban environments, and demographic contexts associated with the identified population groups.

The framework is applied to mobile phone app location data from London and compared with three established methods. ACE supports the identification of six differentiated weekday activity-pattern groups characterised by distinct daily rhythms, urban functional contexts, and demographic associations. The comparative and interpretive analyses demonstrate the analytical utility of the learned representations for revealing behaviourally coherent population heterogeneity from unlabelled mobile phone location data.

The contributions of this study are threefold. Conceptually, we synthesise activity pattern mining from passive location data as an integrated representation–clustering–interpretation framework, clarifying how assumptions made during representation shape downstream pattern discovery and behavioural interpretation. Methodologically, we propose ACE, a self-supervised activity-chain encoder that incorporates pre-trained urban embeddings, visit timing, duration, and sequential organisation without relying on deterministic activity-purpose labels. Empirically, we apply the framework to London mobile phone location data, compare ACE with three established representation methods, and identify six differentiated weekday activity-pattern groups that are interpreted through temporal, functional, and demographic evidence. Together, these contributions establish a holistic route for deriving behaviourally meaningful population patterns from unlabelled mobile phone location data. The source code for the complete analytical pipeline is made publicly available to support reproducibility and further application.


\section{Related work}

\subsection{Activity pattern mining}
As discussed in the Introduction, activity pattern mining generally involves three stages: representation, clustering, and interpretation. Representation converts raw mobility records or activity chains into feature representations; clustering groups individuals with similar patterns; and interpretation analyse the resulting groups. We review prior studies following these three stages.

\subsubsection{Representation}

A fundamental challenge in representing activity patterns from passive mobility data is that activity purpose is typically not directly observed. Researchers have therefore used information about visited places to provide contextual evidence about possible activities. Early approaches commonly derive semantic profiles from land-use or POI information. For example, \cite{cheng2018grouping} used term frequency--inverse document frequency (TF-IDF), while \cite{shen2018profiling} used latent Dirichlet allocation (LDA), to quantify the functional semantics of significant places. Individuals were subsequently represented by their time allocation across these semantic profiles, forming semantic profiles for clustering.

A parallel line of research has focused more strongly on the temporal and sequential structure of mobility. \cite{goulet2016inferring} represented public transport users as multi-week sequences of hourly activity-location states and applied principal component analysis to derive lower-dimensional representations for clustering. \cite{fu2025exploring} subsequently used a stacked denoising autoencoder with deep embedded clustering to model similar multi-week activity-travel sequences. In a survey-based setting, \cite{allahviranloo2017modeling} represented daily activity patterns as time-discretised sequences capturing activity type, duration, and scheduling, and identified representative patterns using sequence-based clustering. These sequence-oriented approaches preserve temporal organisation more explicitly than aggregated semantic profiles, but either rely on labelled activity information from travel surveys or, when applied to smart-card data, represent visited places as user-specific relative location states rather than representations of their functional urban context.


More recent studies have introduced embedding-based representations to capture richer activity information. \cite{li2024multi} inferred non-home and non-work activity types in mobile phone data using a classifier trained on labelled travel-survey data and subsequently treated activity units as words to learn embeddings, which were aggregated into activity-chain representations using Smooth Inverse Frequency. This approach introduces learned semantic similarity between activities, but depends on labelled activity purposes and its aggregation step does not explicitly retain the ordering of stays within the chain. \cite{li2025understanding}, by contrast, represented multimodal trips as textual sequences and used a pre-trained BERT model \citep{devlin2019bert} to obtain sequence-level embeddings. Although this preserves sequence more directly and can be adapted to activity chains, converting mobility observations into textual descriptions may abstract away detailed geographic, temporal, and duration information.

Taken together, existing representation strategies expose several complementary trade-offs. Methods based on semantic profiles can describe the functional composition of visited places but are constrained by predefined or locally derived semantic features and often aggregate activity information. Sequence-oriented approaches preserve temporal organisation but may contain limited information about the functional characteristics of visited environments. More recent embedding-based methods provide richer learned representations, but may depend on labelled activity purposes or abstract mobility observations into predominantly semantic sequences. There remains a need for representation learning methods that can operate when activity-purpose labels are unavailable, encode informative representations of the urban contexts in which stays occur, explicitly incorporate visit timing and duration, and preserve the sequential organisation of daily activity chains.

\subsubsection{Clustering}

Once activity representations have been constructed, clustering is commonly used to identify population groups with similar behavioural patterns. Previous studies have adopted a range of methods, including K-Means, hierarchical clustering, affinity propagation, DBSCAN, and combinations of multiple clustering procedures \citep{goulet2016inferring,shen2016framework,allahviranloo2017modeling,tian2022characterizing,li2024multi,li2025understanding,imran2025novel}. There is no one-size-fits-all approach; rather, the diversity of approaches reflects differences in the structure of the representations, assumptions about cluster geometry, and analytical objectives, rather than the existence of a universally preferred clustering method.


More importantly for activity pattern mining, the behavioural distinctions recoverable by clustering are constrained by the information encoded in the preceding representation stage. Clustering can separate users only along dimensions that the representation preserves; information discarded through semantic simplification, temporal aggregation, or loss of sequence cannot be recovered by selecting a different clustering algorithm. We therefore treat clustering as an essential but methodologically flexible stage and focus primarily on improving the representation on which downstream population grouping depends.

\subsubsection{Interpretation}


The final stage is to interpret and validate the clustering results. Because activity pattern clustering is unsupervised, ground truth labels are rarely available, so evaluation is often qualitative and exploratory. A common approach is to visualise clusters using temporal and semantic heatmaps, showing when different groups tend to conduct different types of activities. This has been widely used to analyse activity-pattern clusters and reveal differences in daily rhythms, activity preferences, and temporal regularities \citep{goulet2016inferring,tian2022characterizing,li2024multi,li2025understanding,imran2025novel}. Further analysis can examine activity duration, trip frequency, spatial extent, visited place types, home area characteristics, and socio-demographic context. 

Interpretation is not merely a post-hoc visualisation step. In the absence of ground-truth activity-pattern labels, it provides the principal means of examining whether clusters correspond to coherent and substantively meaningful forms of behaviour. It also completes the dependency across the three stages: the representations determine which similarities are available to clustering, while interpretation establishes what behavioural meaning can reasonably be attributed to the resulting groups. In this study, temporal and functional patterns are combined with area-level demographic context to characterise the identified clusters and support the development of activity-pattern personas.


\subsection{Pre-trained urban embedding models}




Geospatial representation learning aims to encode spatial and semantic information about geographic entities, areas, or locations into reusable embedding vectors. When learned for urban spaces, these representations are often referred to as urban embeddings \citep{liu2026cityrep}. Recent models use diverse data sources, including remote sensing imagery \citep{klemmer2025satclip,feng2026tessera,brown2025alphaearth}, street view imagery \citep{wang2020urban2vec}, and POIs \citep{huang2022sppe,balsebre2024city,wang2025multi,liu2025enriching,qin2025learning}.

Pre-trained urban embeddings have been applied to socioeconomic mapping, population estimation, environmental modelling \citep{liu2026cityrep}, and human mobility modelling \citep{wang2025into}, demonstrating their ability to provide reusable representations of urban environments across downstream tasks. For activity-pattern analysis, they offer several advantages over place representations derived solely from local POI frequencies or topic models. First, embeddings learned from large-scale geospatial data can encode a broader range of physical and functional characteristics than representations constrained to a selected POI vocabulary or locally estimated topics. Second, pre-training transfers knowledge about urban environments independently of the target mobility dataset, reducing the need to derive place representations from the same observations used for activity pattern modelling. Third, their dense, fixed-dimensional form can be incorporated naturally into neural sequence architectures and self-supervised learning objectives.

\subsection{Persona development}

A persona is a fictional representation of an archetypal user group, originally developed to support human-centred design. In transport research, personas have been used to represent different traveller types, mobility needs, and behavioural patterns, supporting more targeted transport services and policies \citep{vallet2020tangible}.

Recent persona studies increasingly emphasise empirical grounding rather than relying only on expert assumptions \citep{gonzalez2018personas}. Computational segmentation methods have been used to generate personas from observed behavioural data. For example, \citep{stevenson2019personification} developed a computational persona generator to identify representative taxi driver personas in Brazil using survey and individual-level data. This direction is relevant to our study, as mobile phone location data provide large scale observations of daily mobility behaviour.

Activity patterns are also associated with socio-demographic characteristics \citep{zhang2020you}, and recent studies have combined passively observed mobility with contextual demographic information to develop population classifications \citep{mavrogeni2026geodemographic}. Building on these studies, we develop activity personas by combining learned activity pattern representations with socio-demographic features inferred from home locations. The resulting personas are therefore grounded in observed mobility behaviour and interpretable through both activity patterns and demographic context.

\section{Methods}

\begin{figure}
	\centering
	\includegraphics[scale=.95]{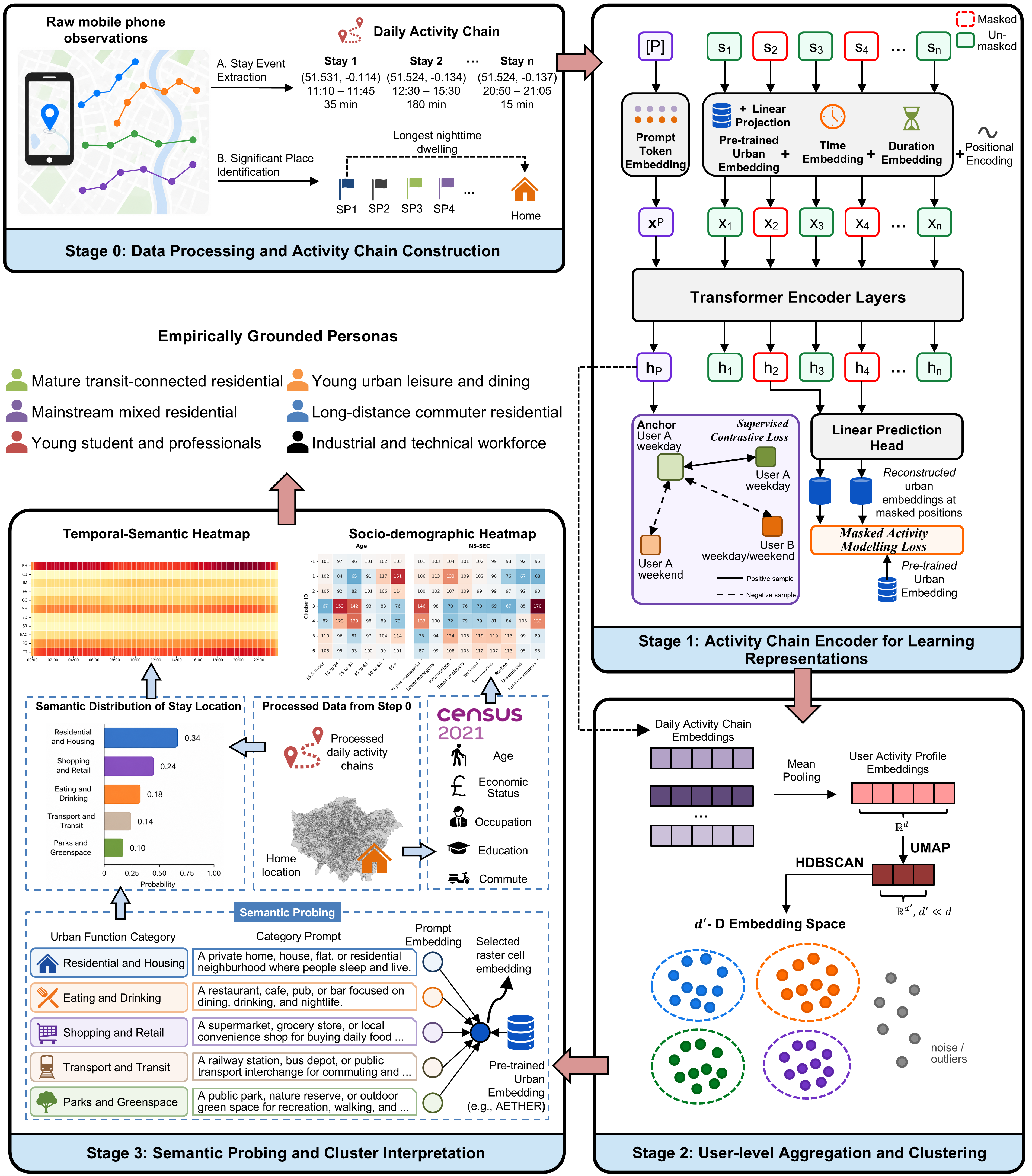}
	\caption{The proposed methodological framework for mining activity patterns from mobile phone data.}
	\label{fig:methodological_framework}
\end{figure}

This study operationalises the representation–clustering–interpretation framework introduced in Section 1. As shown in Figure~\ref{fig:methodological_framework}, the framework comprises a preliminary data processing stage followed by three main modelling and analytical stages. In Stage 0, raw mobile phone observations are processed to extract stay events, identify significant places, and construct daily activity chains. In Stage 1, we introduce the \textbf{Activity Chain Encoder} (\textbf{ACE}), a Transformer-based model that represents each daily activity chain as a sequence of stay tokens and learns contextual chain embeddings through identity-guided contrastive learning and masked activity modelling. In Stage 2, the learned daily activity chain embeddings are aggregated at the user level and clustered to identify population groups with distinct activity-pattern profiles. In Stage 3, semantic probing is used to interpret the spatial context of stay locations, while temporal-semantic and socio-demographic evidence is integrated to characterise the resulting clusters as empirically grounded personas. The preprocessing component is labelled Stage 0 to distinguish it from the three main stages, i.e., representation, clustering, and interpretation.

Let $\mathcal{U}$ denote the set of anonymised users and $\mathcal{D}$ the set of observation days. For each user $u \in \mathcal{U}$ and day $d \in \mathcal{D}$, we construct a daily activity chain 
$
\mathbf{C}_{u,d} =
\left(s_{u,d,1},s_{u,d,2},\ldots,s_{u,d,T_{u,d}}\right),
$
where $s_{u,d,i}$ is the $i$-th stay event and $T_{u,d}$ is the number of stays observed for user $u$ on day $d$. The goal of ACE is to learn a mapping

\begin{equation}
    f_{\theta}: \mathbf{C}_{u,d} \mapsto \mathbf{z}_{u,d} \in \mathbb{R}^{H},
\end{equation}
where $\mathbf{z}_{u,d}$ is a dense representation of the daily activity chain and $H$ is the hidden dimension.

\subsection{Data and activity chain construction}

\subsubsection{Study area and data description}

\begin{figure}
    \centering
    \includegraphics[width=0.5\linewidth]{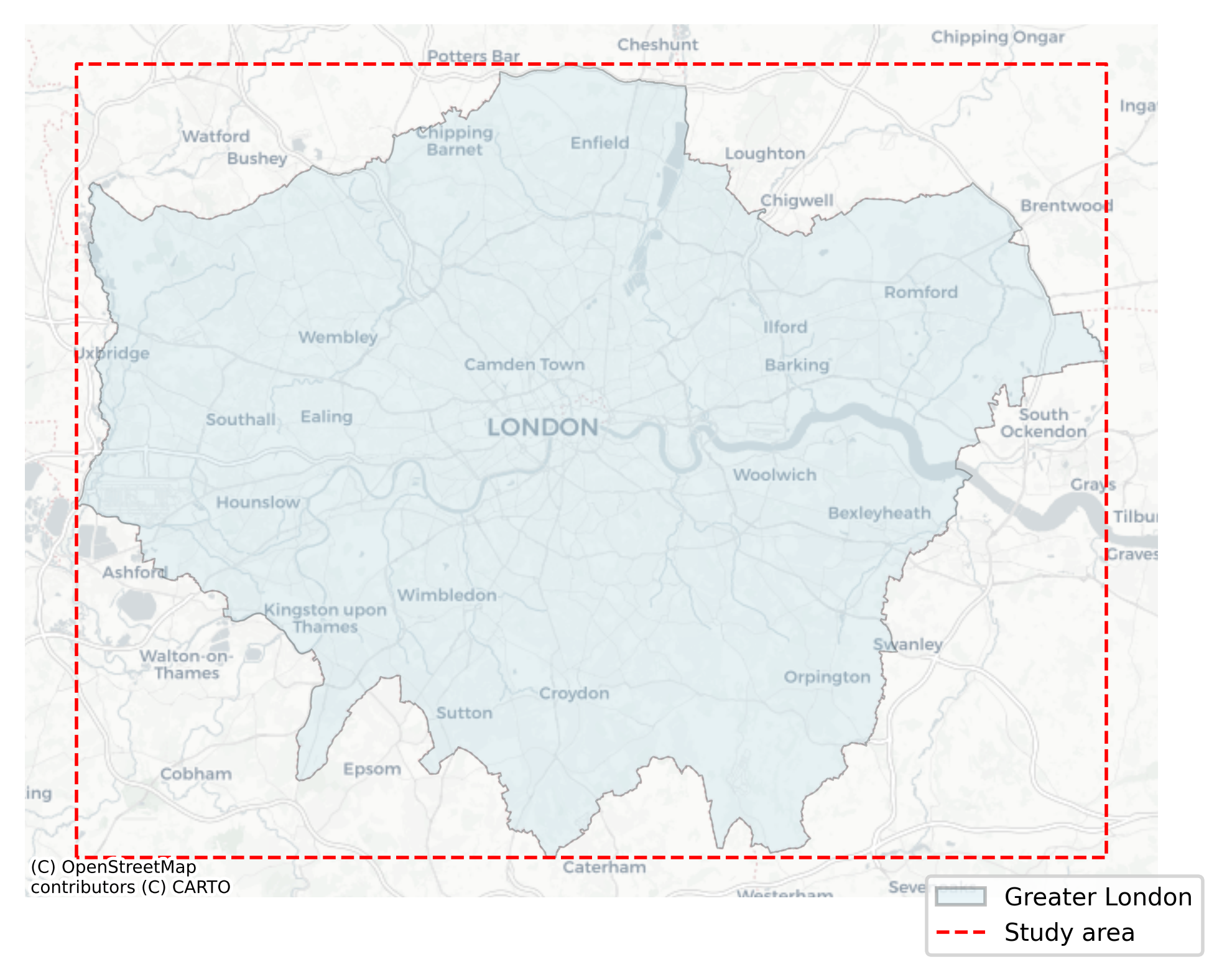}
    \caption{The study area.}
    \label{fig:study_area}
\end{figure}

\paragraph{Study area} The empirical analysis focuses on London. Rather than clipping observations to the exact Greater London administrative boundary, we define the study area using the London bounding box. This avoids truncating cross boundary movements, especially for users who commute into or out of London, and preserves the continuity of daily routines while retaining London as the core spatial context. The administrative boundary and the bounding box used as the study area are shown in Figure \ref{fig:study_area}.

\paragraph{Mobile phone app location data}
The raw mobile phone location data were provided by a commercial data provider and consist of anonymised location traces collected through mobile phone applications. The data used in this study span two full weeks, from 10 to 23 February 2025. In London, the underlying dataset contains approximately 155,000 unique devices per week, corresponding to around 1.8\% of the adult residential population, with a median of 103 observations per user per day and a median horizontal accuracy of approximately 15\,m. 
An internal validation analysis of our data source showed close agreement between the observed devices and Census population distributions across geographic areas.

\paragraph{POI data} The POI data were obtained from Ordnance Survey (OS) via Digimap\footnote{https://digimap.edina.ac.uk/} under an educational licence. The OS POI classification scheme follows a three-tier hierarchy comprising 9 groups, 52 categories, and more than 600 classes\footnote{The detailed classification scheme is available at \href{https://docs.os.uk/os-downloads/products/addresses-and-names-portfolio/points-of-interest/points-of-interest-classification-scheme}{OS Points of Interest Classification Scheme}}. The dataset is regularly maintained and updated, and the March 2025 release was used in this study. After restricting the dataset to POIs located within the study area, 388495 POIs remained, covering 9 unique groups, 52 categories, and 583 classes.

\subsubsection{Stay event detection and attributes construction}

Following common stay detection practice, location points are classified as moving or stationary using speed, acceleration, and spatial and temporal thresholds, after which consecutive stationary records are grouped into stay events. For each stay event, the start and end times are defined by its earliest and latest timestamps, respectively, and the stay duration is calculated as their difference. The stay location is represented by the median coordinate of all records within the segment. Each stay is therefore characterised by its start time, end time, duration, and projected location, as illustrated in Stage 0 of Figure~\ref{fig:methodological_framework}. Formally, a stay can be represented as $s=(t^{start},t^{end},lon,lat)$.

After extracting stay events from the raw data, we performed several additional preprocessing steps. Stays lasting less than 120 seconds are removed to reduce transient and poorly resolved observations. Daily activity chains containing fewer than two stays are also excluded because they provide insufficient sequential information for activity chain modelling. The descriptive statistics of the retained dataset after preprocessing are summarised in Table~\ref{tab:preprocessing_statistics}.

\begin{table}[t]
\centering
\caption{Descriptive statistics of the dataset after preprocessing.}
\label{tab:preprocessing_statistics}
\begin{tabular}{lr}
\toprule
Total number of users                                & 112,112   \\
Total number of stays                                & 3,495,967 \\
Total number of daily activity chains                & 785,476   \\
Mean/median number of daily activity chains per user & 7.0/7     \\
Mean/median length of daily activity chains              & 4.5/4     \\
\bottomrule
\end{tabular}
\end{table}

For each retained stay $s_i$, we derive three discretised temporal and duration attributes: the start-hour bucket $h_i \in \{0,1,\ldots,23\}$, the start-minute bucket $m_i$, and the duration bucket $r_i$. The start-minute bucket indicates the corresponding 15-minute interval within an hour, such that $m_i \in \{0,1,2,3\}$. The duration bucket indicates the corresponding 5-minute interval within a day and is indexed as $r_i \in \{0,1,\ldots,287\}$. Each daily activity chain for user $u$ on day $d$ is additionally assigned a day-of-week indicator $q_{u,d}\in\{0,\ldots,6\}$ (Monday to Sunday) and a binary day-type indicator $w_{u,d}$, where $w_{u,d}=0$ denotes a weekday and $w_{u,d}=1$ denotes a weekend day.


The spatial and semantic context of each stay is represented using pre-trained AETHER urban embeddings \citep{liu2025beyond}. The stay coordinate is mapped to its corresponding raster cell $(row_i,col_i)$, from which a 128-dimensional embedding is retrieved:
$ \mathbf{e}^{loc}_i = \mathbf{E}^\mathrm{AETHER}_{row_i,col_i} \in \mathbb{R}^{128}.$ AETHER was selected for three main reasons. First, it is built upon globally pre-trained AlphaEarth Foundation embeddings, providing broad geographic coverage at a fine spatial resolution of $10 \times 10$ m. Second, it aligns physical characteristics captured by satellite imagery with POI information reflecting socioeconomic and human-centred urban functions, making the resulting representations particularly suitable for urban applications. Third, AETHER is open-source, readily accessible, and has demonstrated strong performance across urban representation benchmarks \citep{liu2026cityrep}. Although ACE is compatible with other pre-trained urban embeddings, AETHER was adopted in this study because of these advantages.

AETHER is used as a frozen spatial-semantic embedding layer within ACE. Its embeddings encode the surrounding urban context, while ACE learns how these contexts are organised into daily behavioural sequences. Freezing the embedding layer also preserves the pre-trained urban representation space and reduces the number of trainable parameters.

\subsubsection{Significant place identification}

Significant places are identified for downstream interpretation and baseline comparison. For each user, stay locations are clustered using DBSCAN in projected coordinates. The resulting clusters are treated as user-specific significant places. The home location is identified as the significant place with the largest accumulated nighttime duration, with total stay duration used as a secondary criterion.

\subsection{Activity chain encoder}

Stage 1, shown in the right-hand panel of Figure~\ref{fig:methodological_framework}, illustrates the architecture and training process of the proposed ACE model. ACE is a sequence representation model for daily stay chains. Its objective is to learn an embedding space in which activity chains with similar behavioural structure are close to one another, while preserving information about where activities occur, when they occur, how long they last, and how they are ordered throughout the day.

Each daily chain, $\mathbf{C}_{u,d}$, is encoded as a sequence of stay tokens. For each stay, ACE combines a frozen urban embedding, trainable temporal embeddings and a trainable duration embedding. The resulting sequence is processed by a Transformer encoder. A learned prompt token representing weekday or weekend day type is prepended to the sequence, and its final hidden state is used as the daily activity chain embedding. The following subsections describe the technical components of ACE in detail.

\subsubsection{Input representation}

For each stay $s_i$, the 128-dimensional AETHER embedding is first projected into the Transformer hidden space:
$
\mathbf{x}^{loc}_i
=
\mathbf{W}^{loc}\mathbf{e}^{loc}_i+\mathbf{b}^{loc},
$
where $\mathbf{W}^{loc}$ and $\mathbf{b}^{loc}$ are trainable parameters of the projection layer. Temporal information is represented using trainable embeddings for hour, start-minute bucket and day of week:
$
\mathbf{x}^{time}_i
=
\mathbf{E}^{hour}_{h_i}
+
\mathbf{E}^{min}_{m_i}
+
\mathbf{E}^{dow}_{q_{u,d}}.
$ And similarly, duration is represented by a learned embedding:
$
\mathbf{x}^{dur}_i
=
\mathbf{E}^{dur}_{r_i}.
$ The stay token representation is obtained by summing these components plus a sinusoidal positional encoding
$\mathbf{p}_i$:

\begin{equation}
    \mathbf{x}_i
    =
    \mathbf{x}^{loc}_i
    +
    \mathbf{x}^{time}_i
    +
    \mathbf{x}^{dur}_i
    +
    \mathbf{p}_i
\end{equation}

To distinguish weekday and weekend routines, ACE prepends a learned day type prompt token:
$
\mathbf{x}^{prompt}_{u,d}
=
\mathbf{E}^{prompt}_{w_{u,d}}
$. The Transformer input sequence is therefore
\begin{equation}
\mathbf{X}_{u,d}
=
\left[
\mathbf{x}^{prompt}_{u,d},
{\mathbf{x}}_1,
{\mathbf{x}}_2,
\ldots,
{\mathbf{x}}_{T_{u,d}}
\right]
\end{equation}

The sequence will be randomly masked before passing through multiple Transformer encoder layers: $
\mathbf{H}_{u,d}
=
\mathrm{TransformerEncoder}_{\theta}(\mathbf{X}_{u,d}),
$
And the daily activity chain embedding is taken as the final hidden state of the prompt token:

\begin{equation}
    \mathbf{z}_{u,d}
    =
    \mathbf{H}^{(0)}_{u,d},
\end{equation}
where the superscript $(0)$ denotes the prompt token’s position at the beginning of the input sequence.

In the main configuration, ACE uses an input hidden dimension of 128, four Transformer encoder layers, eight attention heads, a feed-forward dimension of 256 and dropout of 0.1. We also tested mean pooling over stay token outputs and the concatenation of the prompt token output and the mean-pooled stay token outputs. The prompt-token representation was selected for the main analysis because it produced clustering results that were at least as interpretable as the alternatives while keeping the representation dimension compact.

\subsubsection{Learning objectives}

As shown in Stage 1 of Figure~\ref{fig:methodological_framework}, ACE is trained using a joint objective that combines identity-guided supervised contrastive learning \citep{khosla2020supervised} with masked activity modelling (MAM). 
These objectives are complementary. The masked activity modelling encourages the encoder to learn contextual dependencies among stays within the same chain, while the identity-guided contrastive objective encourages repeated daily routines from the same user and day type to have similar representations.

The masked activity modelling (MAM) objective is inspired by masked language modelling (MLM) in BERT \citep{devlin2019bert}, but differs in the form of the prediction target. In BERT-style MLM, each masked token is predicted through a classification objective over a fixed vocabulary. This formulation is not directly suited to our setting because stay locations are represented by continuous pre-trained AETHER embeddings rather than discrete token identities. Treating all raster cells as a location vocabulary would also lead to a very large candidate space and an expensive output softmax.

We therefore formulate masked stay reconstruction as a contrastive prediction problem over continuous spatial-semantic embeddings, following the broader idea of contrastive masked prediction in self-supervised speech representation learning \citep{baevski2020wav2vec}. Let $\mathcal{M}_i$ denote the set of masked positions in activity chain $i$, obtained by randomly sampling a fixed percentage of its valid input tokens, and let $\mathcal{M}$ be the set of all masked stay positions in a mini-batch. Let $\mathcal{S}$ denote the set of all valid non-padding stay positions in the same mini-batch. For each masked position $a \in \mathcal{M}$, the model predicts a reconstructed location embedding $\hat{\mathbf{e}}^{\mathrm{loc}}_a$ via a linear prediction head. The reconstruction target is the original frozen AETHER embedding $\mathbf{e}^{\mathrm{loc}}_a$.

Each reconstructed embedding is compared with the original AETHER embeddings of
all valid stays in the mini-batch. Both reconstructed and target embeddings are
$L_2$-normalised before similarity is computed. The logits are defined as
\begin{equation}
    \ell_{ab}
    =
    \frac{
    \left(\hat{\mathbf{e}}^{\mathrm{loc}}_a\right)^\top
    \mathbf{e}^{\mathrm{loc}}_b
    }{\tau_\mathrm{MAM}},
    \quad
    a \in \mathcal{M},\ b \in \mathcal{S},
\end{equation}
where $\tau_\mathrm{MAM}$ is the temperature parameter. For each masked stay $a$, the positive
target is its corresponding original embedding, indexed by $b=a$, while the
remaining valid stay embeddings in the mini-batch serve as contrastive candidates.
The MAM loss is then
\begin{equation}
    \mathcal{L}_{\mathrm{MAM}}
    =
    -\frac{1}{|\mathcal{M}|}
    \sum_{a \in \mathcal{M}}
    \log
    \frac{
    \exp(\ell_{aa})
    }{
    \sum_{b \in \mathcal{S}}
    \exp(\ell_{ab})
    } .
\end{equation}
This objective trains ACE to recover the spatial-semantic representation of each masked stay from its surrounding activity context while distinguishing it from other valid stay embeddings in the mini-batch. In doing so, the model learns contextual dependencies among activities within a daily sequence, following a principle analogous to masked representation learning in natural language processing.

Complementing MAM’s token-level objective, identity-guided contrastive learning operates at the level of whole activity chain embeddings. Each daily chain is assigned a unique contrastive label based on the combination of user identity $u$ and day type $w_{u,d}$. In our implementation, string-formatted user identifiers are first encoded as consecutive integers and then combined with the day-type to form unique contrastive classes.

Given a mini-batch of $B$ daily chains, let $\mathbf{z}_i$ denote the $L_2\text{-normalised}$ embedding of chain $i$. For any given anchor chain in the batch, its positive set $P(i)$ consists of all other chains within the same mini-batch that share the exact same contrastive label (i.e., those belonging to the same user and day-type combination).

The supervised contrastive loss is

\begin{equation}
    \mathcal{L}_\mathrm{SC}
    =
    -\frac{1}{B}
    \sum_{i=1}^{B}
    \frac{1}{|P(i)|}
    \sum_{p \in P(i)}
    \log
    \frac{
    \exp(\mathbf{z}_i^\top \mathbf{z}_p / \tau_\mathrm{SC})
    }{
    \sum_{a \neq i}
    \exp(\mathbf{z}_i^\top \mathbf{z}_a / \tau_\mathrm{SC})
    },
\end{equation}

where $\tau_\mathrm{SC}$ is the temperature parameter. This loss pulls together chains from the same user and same day type while separating chains from different user-day-type identities.

The final loss is a weighted sum:

\begin{equation}
    \mathcal{L}
    =
    \mathcal{L}_\mathrm{SC}
    +
    \lambda\mathcal{L}_\mathrm{MAM}.
\end{equation}

In the main configuration, the weighting parameter $\lambda$ is set to 1.0. The two loss values were found to be on a comparable scale, and equal weighting avoids imposing a prior preference for either user-level consistency or within-chain contextual reconstruction.

\subsubsection{Training protocol}

ACE is trained with a user-level train-validation split. Users, rather than individual chains, are partitioned into training and validation sets, which prevents identity leakage across splits. The validation ratio is 0.2.

Mini-batches are constructed using a PK sampling strategy, a common batching scheme in metric learning \citep{hermans2017defense}. In each mini-batch, the sampler first selects $N_{id}$ identities, where each identity is defined by the combination of a user and a day type, and then samples $K_\mathrm{sample}$ daily chains for each identity. This ensures that positive examples are available within each mini-batch for the contrastive objective. In our implementation, the identity is defined as the pair $(u,w)$, where $u$ denotes the user and $w$ denotes the day type, i.e., weekday or weekend. 
We set $K_\mathrm{sample}=2$ so that when multiple chains are available for the same identity, two chains are sampled to form a positive pair. For identities with only one chain available, the chain is duplicated within the batch. Because masking is applied stochastically at runtime, the duplicated chains provide two independently corrupted views of the same daily sequence rather than two fixed identical inputs.

The model is optimised using Adam with a learning rate of 0.001 and batch size 256. The masked activity modelling probability is 0.2. Training is run for a maximum of 50 epochs with early stopping if validation loss does not improve for five consecutive epochs. The checkpoint with the lowest validation loss is retained and used for embedding generation.

\subsection{User representation and clustering}

After ACE training, the learned representations are used to identify groups of users with similar activity-pattern profiles. Clustering is performed on user level embeddings rather than individual daily chain embeddings to characterise and recognise persistent behavioural profiles in the population.

\subsubsection{Aggregation of daily activity chains to form user activity profiles}

ACE produces an embedding $\mathbf{z}_{u,d}$ for each retained user-day activity chain. To obtain activity profile representations at the user level, we apply mean pooling over daily activity chain embeddings.

Let $\eta \in \{\mathrm{weekday}, \mathrm{weekend}\}$ denote the day-type, and let $\mathcal{D}^{\eta}_u$ be the corresponding set of observed days for user $u$. Specifically, $\mathcal{D}^{\mathrm{weekday}}_u$ contains all the weekdays observed for user $u$, while $\mathcal{D}^{\mathrm{weekend}}_u$ corresponds to retained weekend days. The activity profile embedding of user $u$ under day type $\eta$ is computed as

\begin{equation}
\bar{\mathbf{z}}^{\eta}_u
=
\frac{1}{|\mathcal{D}^{\eta}_u|}
\sum_{d \in \mathcal{D}^{\eta}_u}
\mathbf{z}_{u,d}.
\end{equation}

Users are included in a weekday or weekend clustering analysis only if they have at least one day of the corresponding type. In our analysis, weekday and weekend embeddings are clustered separately. This separation is important because weekday and weekend mobility reflect different behavioural mechanisms. Weekday profiles are more closely associated with routine obligations such as commuting and work, whereas weekend profiles may capture more discretionary, leisure and social activities.

\subsubsection{UMAP-HDBSCAN clustering}
\label{sec:method_umap_hdbscan_clustering}

We use a UMAP-HDBSCAN pipeline to cluster the aggregated ACE embeddings. This combination is commonly adopted for clustering high-dimensional representation vectors \citep{grootendorst2022bertopic}. In this pipeline, UMAP first projects the high-dimensional embeddings into a lower-dimensional space while preserving local neighbourhood structure, and HDBSCAN then identifies dense regions in the reduced space as clusters.

This pipeline is appropriate for our task for several reasons. First, the number of behavioural groups is unknown in advance. Second, the embedding space may contain non-spherical structures that are not well represented by centroid-based clustering. Third, some users may exhibit atypical or weakly observed patterns, and HDBSCAN can identify such low-density observations as noise rather than forcing every user into a cluster. Finally, UMAP is fitted using cosine distance, which is aligned with the similarity-based interpretation of ACE embeddings.

For clustering, UMAP is applied to reduce the dimensionality of the ACE embeddings. A two-dimensional UMAP projection is used for visual inspection, while a higher-dimensional UMAP representation is used for HDBSCAN. 


Candidate combinations of \texttt{min\_cluster\_size} and \texttt{min\_samples} are evaluated using three criteria: the number of clusters, the proportion of outlier users, and the density-based clustering validation (DBCV) score. The DBCV score measures the relative density separation and density connectedness of the resulting clusters, with higher values indicating better-defined density structure. Condensed tree diagnostics are also inspected to assess cluster persistence and interpretability. The final parameter setting is selected by jointly considering statistical validity and behavioural interpretability rather than optimising a single metric in isolation.

\subsection{Cluster interpretation}

The clustering stage identifies groups of users with similar learned activity profile embeddings, but the embedding coordinates themselves are not directly interpretable. We therefore develop an interpretation pipeline that links ACE clusters back to the semantic structure of the urban environment, the timing of stays and the socio-demographic context of users' inferred home areas. The interpretation stage has four components: semantic probing of the pre-trained urban embeddings, assignment of semantic distributions to stay events, temporal-semantic heatmap construction and persona profiling.

\subsubsection{Semantic probing of urban embeddings}
\label{sec:method_semantic_probing}

\begin{table}
\centering
\caption{Semantic probing prompts used to interpret the urban functional semantics of places.}
\label{tab:probing_prompts}
\small
\begin{tabularx}{\textwidth}{@{}p{0.25\textwidth}X@{}}
\toprule
\textbf{Urban function} & \textbf{Probing prompt} \\
\midrule
Residential and Housing (RH)
& A private home, house, flat, or residential neighbourhood where people sleep and live. \\

Commercial and Business (CB)
& A corporate office, workplace, or business district for daytime professional employees. \\

Industrial and Manufacturing (IM)
& An industrial estate, factory, warehouse, or manufacturing plant with heavy machinery. \\

Education and Schools (ES)
& An academic classroom, school, or university campus during active daytime teaching hours. \\

Grocery and Convenience (GC)
& A supermarket, grocery store, or local convenience shop for buying daily food and household items. \\

Medical and Healthcare (MH)
& A hospital or healthcare centre providing medical treatment, ambulance, and emergency services. \\

Eating and Drinking (ED)
& A restaurant, cafe, pub, or bar focused on dining, drinking, and nightlife. \\

Shopping and Retail (SR)
& A high street, shopping centre, or retail store selling clothes, electronics, and consumer goods. \\

Entertainment, Arts and Culture (EAC)
& A cinema, theatre, museum, or venue for arts, culture, and live entertainment. \\

Parks and Greenspace (PG)
& A public park, nature reserve, or outdoor green space for recreation, walking, and nature. \\

Transport and Transit (TT)
& A railway station, bus depot, or public transport interchange for commuting and travel. \\
\bottomrule
\end{tabularx}
\end{table}

ACE uses pre-trained AETHER embeddings \citep{liu2025beyond} as spatial-semantic inputs. These embeddings encode information about the urban environment, but they are continuous latent vectors rather than explicit land use or activity labels. To interpret the learned clusters, we first conduct semantic probing of the AETHER embedding space with natural language prompts and convert each raster cell into a soft distribution over predefined urban semantic categories.

Let $\mathcal{G}=\{g_1,\ldots,g_K\}$ denote the semantic category set, where $K=11$. The categories are Residential and Housing (RH), Commercial and Business (CB), Industrial and Manufacturing (IM), Education and Schools (ES), Grocery and Convenience (GC), Medical and Healthcare (MH), Eating and Drinking (ED), Shopping and Retail (SR), Entertainment, Arts and Culture (EAC), Parks and Greenspace (PG), and Transport and Transit (TT). These categories were developed with reference to the classification scheme of the original Ordnance Survey Points of Interest data and consolidated into broader, interpretable groups covering major urban functions and everyday activity settings. Similar categories are commonly used in POI-based urban studies \citep{zhang2017hierarchical,niu2023understanding}. However, the category set, together with the corresponding prompts, are not fixed and may be customised, to suit specific urban contexts and research objectives. Each category is represented by a descriptive probing prompt rather than a short label alone, so that the text encoder receives richer contextual information. The full set of probing prompts is reported in Table \ref{tab:probing_prompts}.

For category $g_k$, let $\mathbf{a}_k$ be the raw text embedding of its probing prompt derived from the language model (text encoder) that was used to train AETHER. The trained AETHER text projection head maps this vector into the shared urban embedding space:

\begin{equation}
    \mathbf{q}_k
    =    
    \phi_{\psi}(\mathbf{a}_k),
\end{equation}
where $\phi_{\psi}(\cdot)$ is the learned text projection. Let $\mathbf{e}_{r,c}$ be the AETHER embedding at raster cell $(r,c)$. Both $\mathbf{q}_k$ and $\mathbf{e}_{r,c}$ are $L_2$-normalised before their similarity is computed:

\begin{equation}
    \rho_{k,r,c}
    =
    \mathbf{q}_k^\top \mathbf{e}_{r,c}.
\end{equation}

The similarity scores are converted into a relative semantic distribution using a temperature-scaled softmax:

\begin{equation}
    \pi_{k,r,c}
    =    
    \frac{
    \exp(\rho_{k,r,c}/\tau_s)
    }{
    \sum_{\ell=1}^{K}
    \exp(\rho_{\ell,r,c}/\tau_s)
    },
    \qquad
    k=1,\ldots,K.
\end{equation}
The temperature parameter $\tau_s$ equals to the one used in AETHER pre-training.
The semantic distribution vector characterises the semantic meaning that each raster cell possesses, which can be written as:
\begin{equation}
    \boldsymbol{\pi}_{r,c}
    =
    (\pi_{1,r,c},\ldots,\pi_{K,r,c}),
    \qquad
    \sum_{k=1}^{K}\pi_{k,r,c}=1.
\end{equation}
This soft representation is preferable to hard classification because many urban locations are functionally mixed. For example, a transport hub may also contain retail, eating and drinking, and commercial functions. We use the full semantic distribution to preserve this ambiguity. For readers’ reference, the highest-scoring functional category (i.e., the argmax category) of each raster cell is mapped in Appendix Figure \ref{app_fig:urban_functions}.



\subsubsection{Construction of temporal-semantic heatmaps}

\begin{figure}
    \centering
    \includegraphics[width=0.75\linewidth]{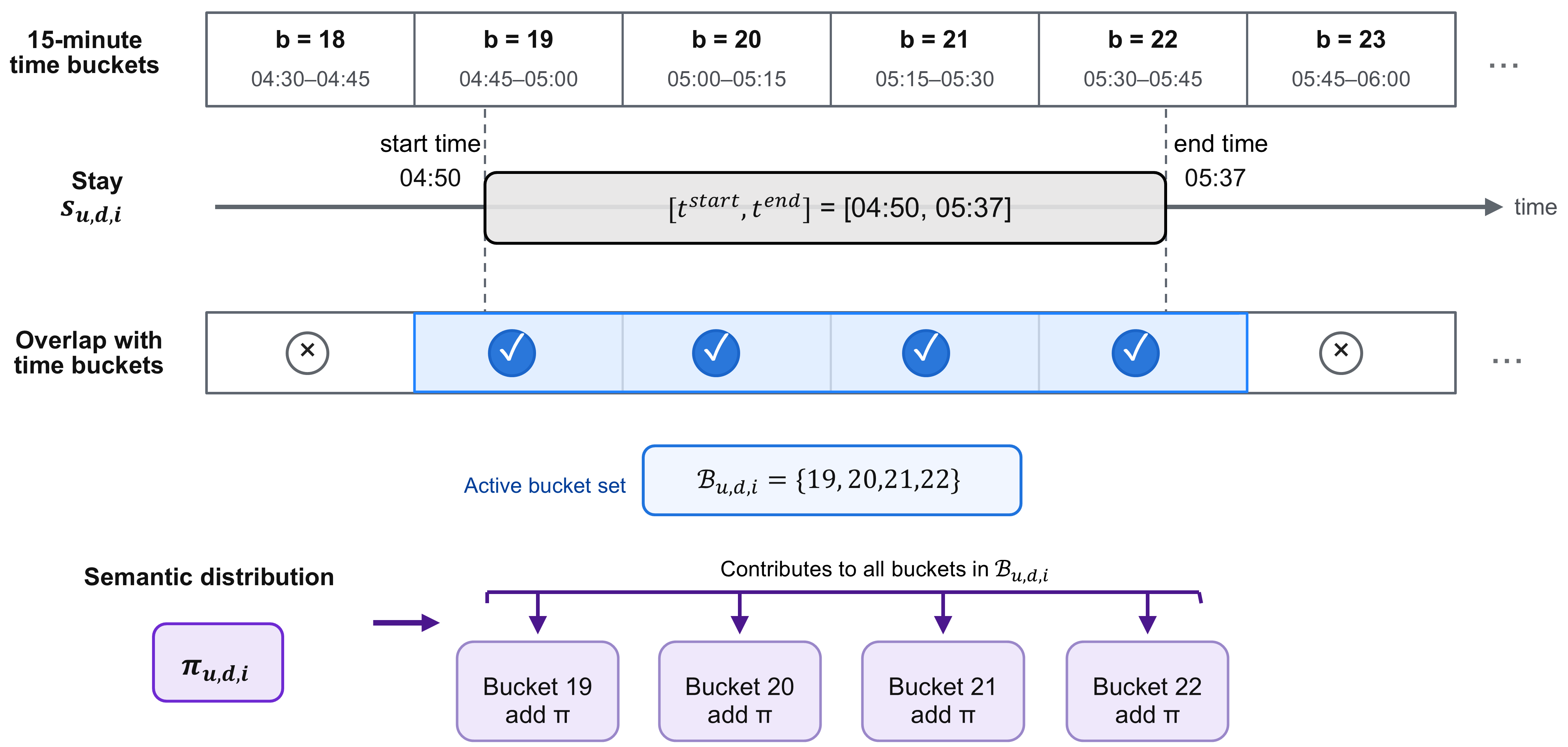}
    \caption{Illustration of the allocation of a stay's semantic probability
    vector to 15-minute time buckets.}
    \label{fig:time_overlap}
\end{figure}

To characterise the temporal and semantic structure of the inferred activity
patterns, each stay is first assigned a semantic distribution according to the
semantic raster cell containing its location. Specifically, stay $s_{u,d,i}$ is
mapped to raster cell
$({row}_{u,d,i},col_{u,d,i})$, from which it inherits the
corresponding semantic probability vector:
$\boldsymbol{\pi}_{u,d,i}
=
\boldsymbol{\pi}_{{row}_{u,d,i},{col}_{u,d,i}}$. Because this representation is probabilistic rather than deterministic, a stay
may contribute to multiple semantic categories simultaneously.

As illustrated in Figure~\ref{fig:time_overlap}, each day is divided into 96
consecutive 15-minute buckets. Let $b \in \{0,\ldots,95\}$ index these buckets,
and let $\mathcal{B}_{u,d,i}$ denote the set of buckets whose intervals have a
non-zero temporal overlap with stay $s_{u,d,i}$. The stay contributes its
semantic probability vector to every bucket in $\mathcal{B}_{u,d,i}$.
Consequently, longer stays generally contribute to more time buckets, allowing
duration information to be incorporated naturally into the temporal-semantic
representation.

Temporal-semantic heatmaps are constructed separately for the weekday and
weekend clustering results. Recall that
$\eta \in \{\mathrm{weekday},\mathrm{weekend}\}$ denotes the day type and that
$\mathcal{D}^{\eta}_u$ is the set of observed days of type $\eta$ for user $u$.
Let $\mathbf{A}^{\eta}_u \in \mathbb{R}^{96 \times K}$ denote the user-level
temporal-semantic matrix, where $K$ is the number of semantic categories. Its
$(b,k)$-th entry is defined as

\begin{equation}
A^{\eta}_{u,b,k}
=
\frac{1}{|\mathcal{D}^{\eta}_u|}
\sum_{d \in \mathcal{D}^{\eta}_u}
\sum_{i=1}^{T_{u,d}}
\mathbb{I}\!\left(b \in \mathcal{B}_{u,d,i}\right)
{\pi}_{u,d,i,k},
\end{equation}

where $T_{u,d}$ is the number of stays in the activity chain of user $u$ on day
$d$, and $\pi_{u,d,i,k}$ is the probability assigned to semantic category $k$
for stay $i$. The resulting matrix represents the user's average daily
temporal-semantic profile. Averaging across observed days at the user level
prevents users with more recorded days from contributing disproportionately to
the cluster profile.

Each user-level matrix is then min--max normalised jointly across all time
buckets and semantic categories, yielding
$\widetilde{\mathbf{A}}^{\eta}_u$. For cluster $c$, the temporal-semantic
heatmap is obtained by averaging the normalised matrices of its members:

\begin{equation}
\bar{A}^{\eta}_{c,b,k}
=
\frac{1}{|\mathcal{U}^{\eta}_{c}|}
\sum_{u \in \mathcal{U}^{\eta}_{c}}
\widetilde{A}^{\eta}_{u,b,k},
\end{equation}

where $\mathcal{U}^{\eta}_{c}$ denotes the set of users assigned to cluster $c$
in the corresponding weekday or weekend analysis. This aggregation gives each
user equal weight in the cluster-level profile.

To highlight how each cluster differs from the overall population, we also
construct a deviation heatmap. The population baseline is calculated by
averaging the normalised matrices of all users included in the corresponding
analysis:

\begin{equation}
\bar{A}^{0,\eta}_{b,k}
=
\frac{1}{|\mathcal{U}^{\eta}|}
\sum_{u \in \mathcal{U}^{\eta}}
\widetilde{A}^{\eta}_{u,b,k},
\end{equation}

where $\mathcal{U}^{\eta}$ denotes the set of all users included in the
weekday or weekend analysis. The deviation heatmap is then defined as

\begin{equation}
\Delta^{\eta}_{c,b,k}
=
\bar{A}^{\eta}_{c,b,k}
-
\bar{A}^{0,\eta}_{b,k}.
\end{equation}

Positive values indicate semantic categories that are over-represented in a
cluster at a given time, whereas negative values indicate under-representation
relative to the corresponding population baseline.

\subsubsection{Socio-demographic contextualisation of clusters}

Finally, the activity-pattern clusters are linked to residential geography and
area-level socio-demographic context. The inferred home location of each user is
spatially joined to a Lower-layer Super Output Area (LSOA), through which the
corresponding Census distributions are assigned to that user. 

Five demographic dimensions are considered: age, National Statistics Socio-economic Classification (NS-SEC), occupation, education level, and commute distance. These variables have been used in previous research to characterise socio-demographic and commuting differences associated with activity patterns \citep{goulet2016inferring,ahmed2021microscopic}. However, the framework is flexible, and other demographic variables may be incorporated depending on the research context and data availability. Detailed definitions of the categories within each dimension are provided in Appendix Table~\ref{tab:demographic_categories}.

For demographic variable $v$, corresponding to one of the aforementioned dimensions, let
$\mathbf{d}^{(v)}_{l}$ denote the vector of category proportions for LSOA $l$,
and let $l_u$ denote the inferred home LSOA of user $u$. For cluster $c$ in the
day-type analysis $\eta$, the cluster-level demographic profile is calculated
as

\begin{equation}
\mathbf{D}^{(v),\eta}_c
=
\frac{1}{|\mathcal{U}^{\eta}_c|}
\sum_{u \in \mathcal{U}^{\eta}_c}
\mathbf{d}^{(v)}_{l_u},
\end{equation}

where $\mathcal{U}^{\eta}_c$ is the set of users assigned to cluster $c$ in the
corresponding weekday or weekend analysis. This formulation gives each user
equal weight; consequently, an LSOA contributes in proportion to the number of
cluster members whose inferred homes are located within it.

Let $\mathbf{D}^{(v)}_0$ denote the reference profile, defined as the London-wide Census distribution for demographic variable $v$. Within each variable, $m$ denotes a specific category, such as the 25–34 age group or professional occupations. The index score for category $m$ is computed as

\begin{equation}
I^{(v),\eta}_{c,m}
=
100
\frac{D^{(v),\eta}_{c,m}}
     {D^{(v)}_{0,m}}.
\end{equation}

An index score of 100 indicates parity with the London-wide reference profile,
whereas values above and below 100 indicate over- and under-representation,
respectively. Similar index scores have been used in previous studies to
facilitate the interpretation of cluster characteristics
\citep{liu2021identifying}.

Because demographic attributes are linked through the area-level context of inferred home locations rather than observed at the individual level, the socio-demographic profiles are interpreted as ecological contextualisation rather than individual-level inference.

The temporal-semantic heatmaps and area-level socio-demographic profiles provide complementary perspectives on each cluster. By jointly interpreting its characteristic activity timing, urban functional context, and residential socio-demographic context, we construct an integrated pen portrait for each cluster. These pen portraits constitute the final persona results, summarising the distinctive behavioural and contextual characteristics associated with each inferred activity-pattern group.

\subsection{Baseline methods and comparison protocol}
\label{sec:baseline_methods}

\begin{figure}
    \centering
    \includegraphics[width=0.8\linewidth]{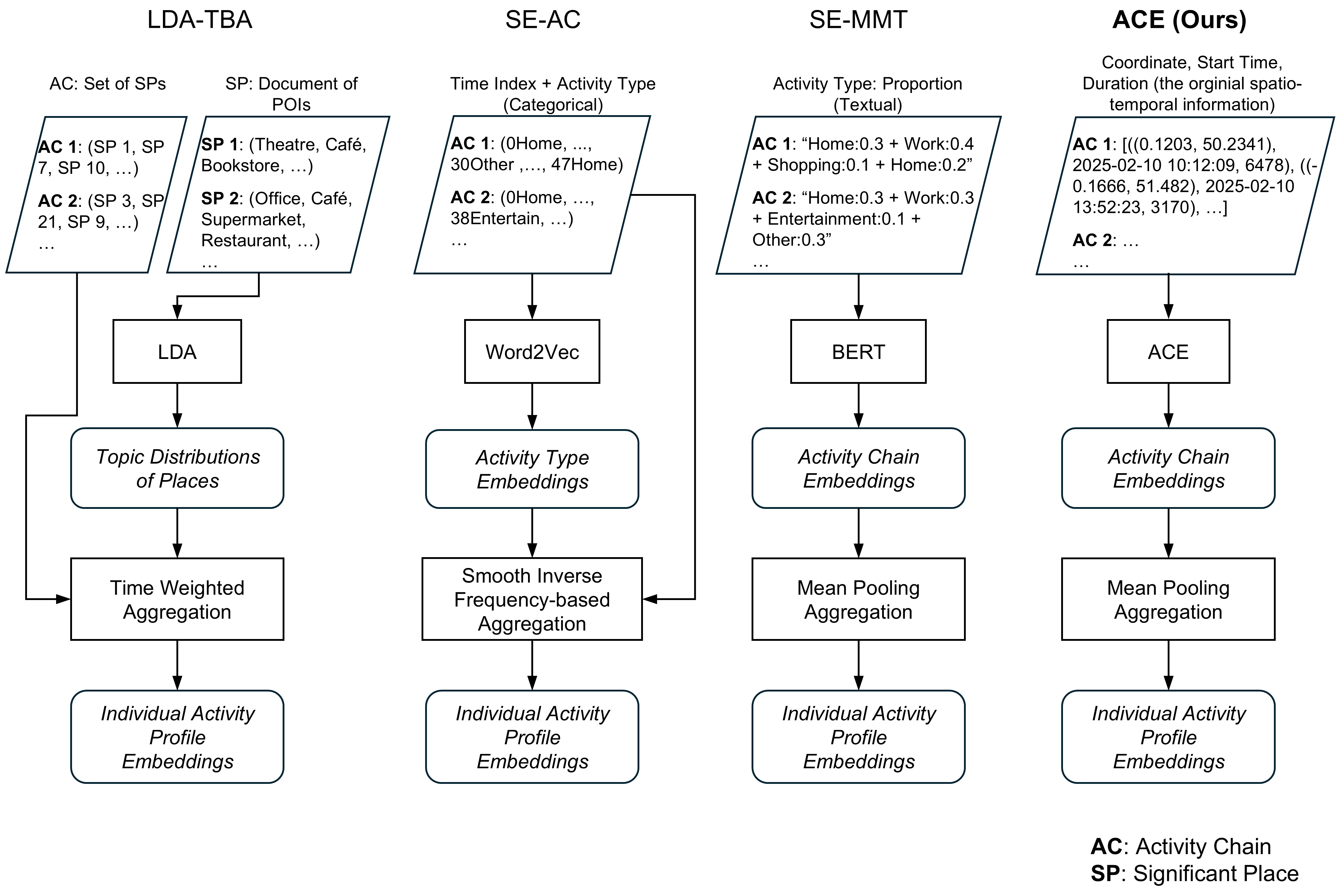}
    \caption{Comparison of the representation stages of LDA-Time Budget Allocation (LDA-TBA), Semantic Embedding of Activity Chains (SE-AC), Semantic Embedding of Multimodal Trips (SE-MMT), and the proposed Activity Chain Encoder (ACE). The methods differ in how activity-chain information is encoded and aggregated into individual activity-profile embeddings. AC denotes activity chain, and SP denotes significant place.}
    \label{fig:comp_baselines}
\end{figure}

To provide additional empirical evidence for the effectiveness of ACE, we compare it with three representative activity pattern representation methods: LDA-Time Budget Allocation (LDA-TBA) \citep{cheng2018grouping,shen2018profiling}, Semantic Embedding of Activity Chains (SE-AC) \citep{li2024multi}, and Semantic Embedding of Multimodal Trips (SE-MMT) \citep{li2025understanding}. These methods represent three different approaches to activity pattern representation: topic-based semantic profiling, activity-type embedding, and language-model-based sequence embedding, respectively. Their representation workflows are summarised in Figure~\ref{fig:comp_baselines}.

Both SE-AC and SE-MMT were originally designed for mobility records with activity-purpose labels. Because such labels are unavailable in the mobile phone data used here, an activity type is inferred for each stay from its surrounding POIs. Rather than assigning the class of the nearest POI, which would assume that spatial proximity alone identifies the activity being undertaken, we use a TF-IDF-based procedure previously adopted for semantic place profiling \citep{cheng2018grouping,shen2018profiling}. TF-IDF scores are calculated for POI classes around each stay, and the class with the highest score is assigned as its deterministic activity label. The classes correspond to the most granular level of the Ordnance Survey POI classification scheme.

For SE-AC, the resulting activity-label sequences are used to train activity-type embeddings using the continuous bag-of-words variant of Word2Vec. The embeddings within each activity chain are then combined using Smooth Inverse Frequency weighting to obtain a chain-level representation. For SE-MMT, each activity chain is converted into a semantic sentence consisting of \texttt{activity\_type:proportion} entries separated by ``+''. The sentence is encoded using the pre-trained \texttt{bert-base-uncased} model, following the original study, and average pooling is applied to the final hidden states to obtain an activity-chain embedding. The chain-level representations produced by both methods are subsequently aggregated at the individual level.

LDA-TBA does not require deterministic activity labels. Instead, LDA is applied to the POIs surrounding each significant place to derive a semantic topic distribution, and an individual profile is constructed by aggregating the distributions of visited places according to the time spent at each location.

The resulting individual-level representations are processed using the common dimensionality-reduction and clustering method described in Section \ref{sec:method_umap_hdbscan_clustering}. The same UMAP hyperparameters are applied to all methods, while the HDBSCAN hyperparameters are optimised separately for each representation to accommodate differences in their embedding structures. This design maintains a consistent downstream pipeline while allowing an appropriate clustering configuration for each method. The comparison therefore primarily reflects differences among the representation approaches rather than arbitrary differences in the downstream procedure.



\section{Empirical results}
\label{sec:results}

This section presents the empirical results of weekday activity patterns in four stages. First, we describe the clustering of individual activity profile embeddings and the relative sizes of the identified clusters. Second, we examine the cluster-level temporal-semantic activity patterns, including both their absolute visitation intensities and their deviations from the population-wide baseline. Third, we compare the demographic contexts associated with the clusters using Census-derived index scores. Finally, the behavioural and demographic evidence is integrated to construct an interpretable typology of weekday activity patterns.

\subsection{Activity patterns identified from ACE representations}
\label{sec:ace_results}

\subsubsection{Activity profile clustering and cluster composition}
\label{subsec:clustering_results}

The individual activity profile embeddings were constructed by mean-pooling the daily activity chain embeddings associated with each individual. The resulting representations were reduced to 5 dimensions using UMAP and subsequently clustered using HDBSCAN. The clustering procedure identified six clusters, together with a set of observations classified as noise. Of the 104711 individuals included in the analysis, 67948 were assigned to one of the six clusters, while 35.11\% of the sample, were classified as noise.


The HDBSCAN clustering was performed in the 5-dimensional reduced embedding space. The results reveal substantial variation in cluster size. Cluster 2 is the largest group, accounting for approximately 44.6\% of clustered individuals, followed by Cluster 5 with approximately 25.9\%. The remaining clusters are smaller and account for between approximately 4.9\% and 11.0\% of the clustered population. The uneven cluster sizes suggest that a large proportion of individuals share relatively common weekday activity profiles, while several smaller groups exhibit more distinctive patterns. 





\subsubsection{Cluster-level temporal-semantic activity patterns}
\label{subsec:activity_patterns}

\begin{figure}
    \centering
    \includegraphics[width=\linewidth]{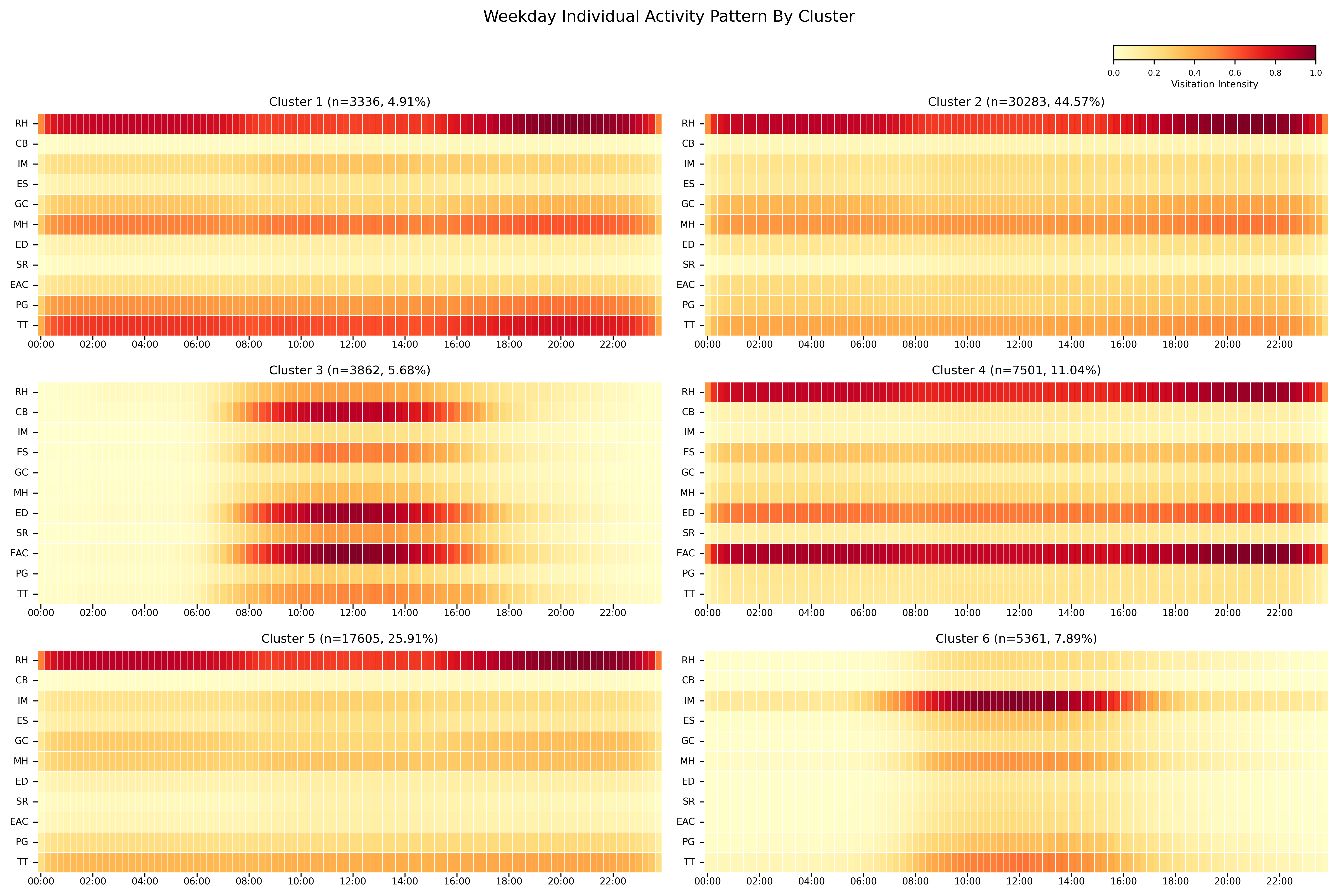}
    \caption{Cluster-level weekday temporal-semantic activity patterns. Each panel shows the mean visitation intensity across hourly intervals and urban functional dimensions for individuals assigned to the corresponding cluster. For the full definition of the semantic categories, see Table \ref{tab:probing_prompts}.}
    \label{fig:absolute_weekday_activity_patterns}
\end{figure}

\begin{figure}
    \centering
    \includegraphics[width=\linewidth]{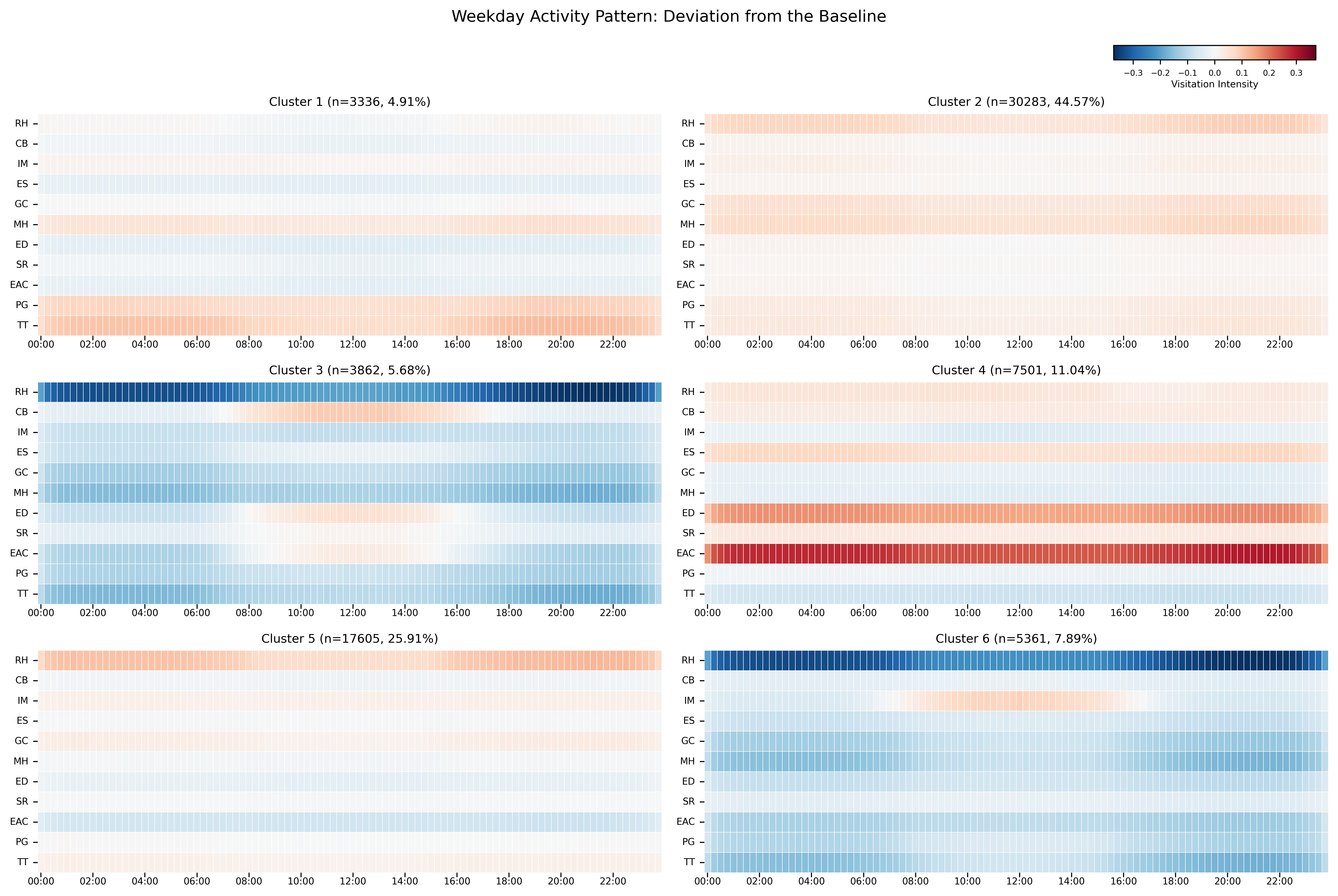}
    \caption{Deviation of each cluster-level weekday temporal-semantic activity pattern from the population-wide baseline. Warm colours indicate above-baseline visitation intensity, while cool colours indicate below-baseline intensity.}
    \label{fig:weekday_activity_deviation}
\end{figure}

To interpret the behavioural structure captured by the learned representations, cluster-level temporal-semantic activity patterns were calculated by averaging the individual profiles of all members assigned to each cluster. Each heatmap represents visitation intensity across hourly intervals and eleven urban functional dimensions defined in Section \ref{sec:method_semantic_probing}.

Figure \ref{fig:absolute_weekday_activity_patterns} presents the absolute weekday temporal-semantic patterns. To distinguish cluster-specific characteristics from activity components that are common across the population, a population-wide baseline was additionally calculated by averaging the weekday profiles of all individuals. Figure \ref{fig:weekday_activity_deviation} shows the deviation of each cluster from this baseline, where positive values indicate above-average visitation intensity and negative values indicate below-average intensity. The absolute heatmaps therefore describe the overall composition of each pattern, while the deviation heatmaps reveal the semantic-temporal features that most clearly differentiate each cluster.

Two broad temporal structures are evident. Clusters 1, 2, 4 and 5 are comparatively residentially anchored and maintain activity throughout the day. In contrast, clusters 3 and 6 exhibit concentrated daytime patterns with substantially lower residential intensity, suggesting regular observed presence in non-residential daytime environments.

The clusters also differ in their functional specialisation. Cluster 2 remains closest to the population-wide baseline and represents a broadly mixed residential pattern. Cluster 1 is distinguished by above-average Transport and Transit, Medical and Healthcare, and Parks and Greenspace intensity. Cluster 4 shows particularly strong Entertainment, Arts and Culture and Eating and Drinking components, especially during the evening. Cluster 5 is characterised by residential dominance and comparatively weak exposure to commercial, cultural and leisure functions. Among the daytime-oriented clusters, Cluster 3 is associated with commercial, dining, cultural and educational environments, whereas Cluster 6 is primarily distinguished by Industrial and Manufacturing activity during conventional working hours.

These semantic dimensions describe the functional contexts of visited locations rather than directly observed activity purposes. They should therefore be interpreted as temporal--semantic activity signatures, rather than definitive accounts of individuals' motivations.

\subsubsection{Demographic characteristics associated with the clusters}
\label{subsec:demographic_results}

\begin{figure}
    \centering
    \includegraphics[width=\linewidth]{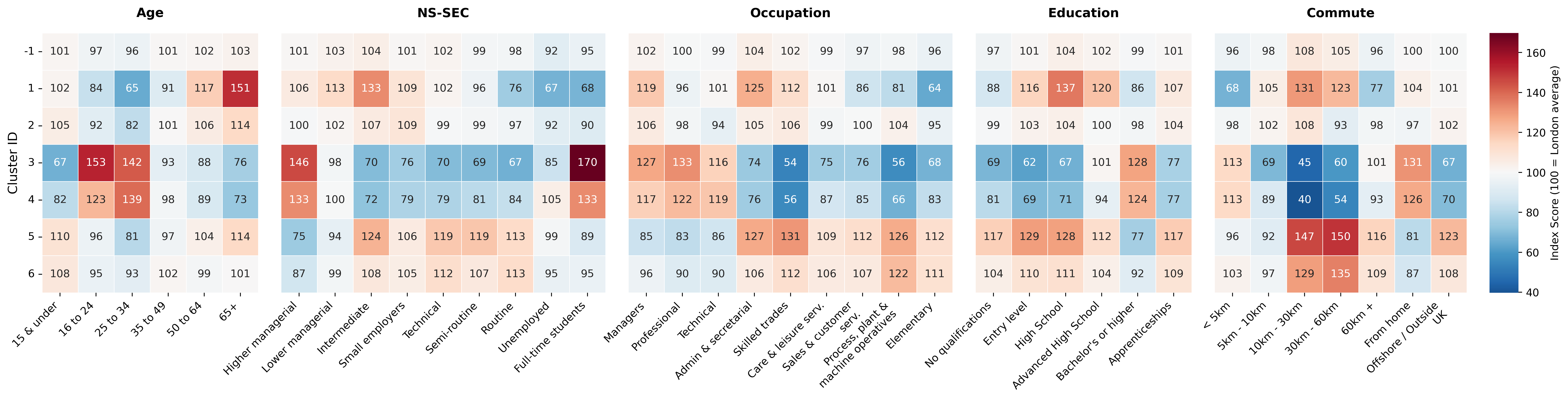}
    \caption{Demographic index scores associated with the inferred home locations of individuals in each activity-profile cluster. An index score of 100 indicates parity with the London-wide reference distribution.}
    \label{fig:demographic_heatmap}
\end{figure}

Figure \ref{fig:demographic_heatmap} compares the demographic contexts associated with the inferred home locations of individuals in each cluster. Five dimensions are considered: age, National Statistics Socio-economic Classification, occupation, education level, and commute distance (see Appendix Table \ref{tab:demographic_categories} for the full attribute details). Index scores are expressed relative to the London-wide Census distribution, with 100 indicating parity. The demographic index scores for both the six clusters and the noise (outlier) group are presented.

The clusters form three broad demographic groupings. Clusters 3 and 4 are associated with younger, highly educated and professionally oriented residential contexts. Both over-represent people aged 16--34, full-time students, managerial and professional socioeconomic groups, degree-level qualifications, short commutes and working mainly from home. Cluster 3 has the stronger student signature, while Cluster 4 is particularly associated with young professional and degree-qualified populations.

Clusters 5 and 6 are associated with more technical, operational and routine employment contexts. Skilled trades, process, plant and machine operatives, lower supervisory and technical occupations, and lower or vocational qualification levels are comparatively prominent. Medium- and long-distance commuting are also over-represented, while home working is less common. These characteristics are strongest in Cluster 5 for commuting distance and in Cluster 6 for industrial and operative occupations.

Clusters 1 and 2 are more residentially mainstream. Cluster 1 is associated with an older population and with intermediate, administrative and moderately skilled occupational groups. Cluster 2 remains close to the London-wide reference across most demographic categories, consistent with its comparatively general activity pattern.

The demographic profile of the noise group is generally close to the London-wide reference across all five dimensions. Most index scores are close to 100, and no demographic category is strongly over- or under-represented. This indicates that the outlier group does not correspond to a single, demographically distinctive population. Together with its classification as noise by HDBSCAN, this finding supports the interpretation that the method separates prominent and internally coherent activity-pattern groups from individuals whose activity-profile embeddings are more heterogeneous and do not form sufficiently dense clusters.


\subsubsection{Integrated activity pattern pen portraits}
\label{subsec:integrated_typology}



The temporal-semantic and demographic results provide complementary forms of evidence. The activity heatmaps reveal when and in what types of urban environments individuals are observed, while the demographic index scores describe the population contexts associated with their inferred residential locations. Integrating these dimensions results in six interpretable pen portraits summarised below.


\paragraph{Cluster 1: Mature transit-connected residential pattern.}
It is residentially anchored but over-represents transport, healthcare and greenspace environments. Its associated home areas contain comparatively older populations and more intermediate, administrative and moderately skilled occupational groups. The cluster therefore reflects a mature residential pattern with strong connections to transport, healthcare and green-space contexts.

\paragraph{Cluster 2: Mainstream mixed residential pattern.}
Cluster 2 is the largest group and its temporal-semantic and demographic profiles are both close to the London-wide baseline. It represents a broadly mixed residential routine without strong functional or demographic specialisation and provides a useful reference against which the remaining clusters can be compared.

\paragraph{Cluster 3: Young student and professional daytime pattern.}
Cluster 3 exhibits a concentrated daytime pattern associated with commercial, dining, cultural and educational environments. Its home areas strongly over-represent young adults, full-time students, professionals and degree-qualified residents, together with short commutes and home working. It therefore captures a young, highly educated population oriented towards active daytime urban destinations.

\paragraph{Cluster 4: Young urban leisure and dining pattern.}
Cluster 4 combines a residential anchor with strong entertainment, cultural and eating and drinking activity, particularly during the evening. The associated demographic context is young, highly educated and professionally oriented. The cluster reflects an amenity-rich urban lifestyle pattern centred on residential, cultural and dining environments.

\paragraph{Cluster 5: Long-distance commuter residential pattern.}
Cluster 5's activity profile is strongly residential and comparatively low in functional diversity. Its associated home areas over-represent technical, administrative, routine and operational occupations, lower or vocational qualifications, and medium- to long-distance commuting. It therefore reflects a residentially anchored routine structured around longer journeys to relatively fixed workplaces.

\paragraph{Cluster 6: Industrial and technical workforce pattern.}
Cluster 6 displays a concentrated daytime Industrial and Manufacturing signature. Its associated home areas over-represent process, plant and machine operatives, skilled trades, technical and routine socioeconomic groups, vocational qualifications and medium- to long-distance commuting. The close correspondence between its activity and demographic profiles supports its interpretation as an industrial and technical workforce pattern.

Overall, the six pen portraits distinguish residentially anchored routines from daytime destination-oriented routines, while also revealing differences in functional context and socioeconomic composition. The correspondence between the independently derived temporal-semantic and demographic profiles indicates that the learned activity-profile embeddings capture meaningful variations in weekday activity behaviour.

\subsection{Comparison with baseline representations}
Having examined the activity-pattern profiles identified by ACE, we now compare its clustering outcomes with those obtained using LDA-TBA, SE-AC, and SE-MMT. 
The resulting clusters are examined using the same temporal-semantic interpretative method applied to ACE, focusing on their behavioural coherence, distinctiveness, and interpretability.

\subsubsection{LDA-TBA}

\begin{figure}
    \centering
    \includegraphics[width=\linewidth]{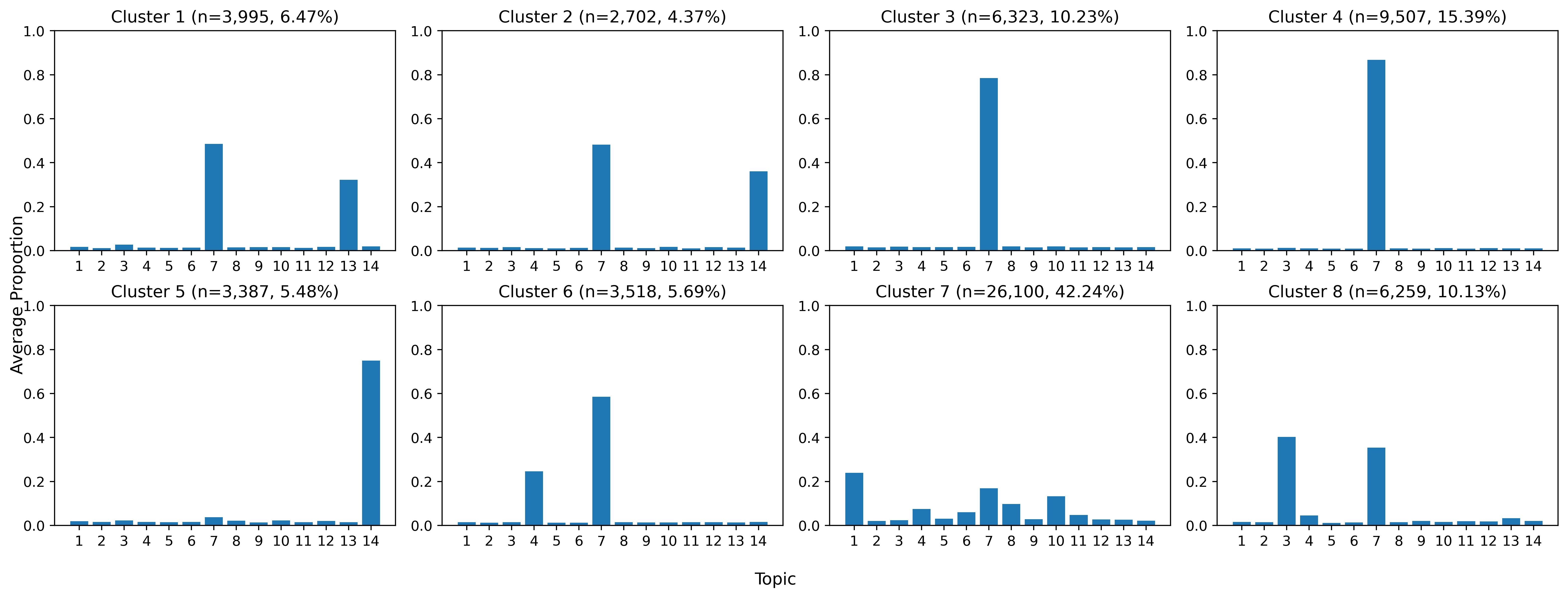}
    \caption{Cluster-level topic distributions derived using LDA-TBA. Each panel shows the average proportion of the 14 LDA topics within a cluster, with cluster size and population share reported in the panel title. The topics were selected using perplexity and are defined by their highest-weighted POI classes in Table \ref{tab:lda_topics}.}
    \label{fig:lda_tba_clusters}
\end{figure}

The number of LDA topics was determined based on perplexity. The identified 14 topics and their highest-weighted POI classes are reported in Table \ref{tab:lda_topics} in the Appendix. Several frequently occurring classes, including bus stops, restaurants, cafes, convenience stores, and hair and beauty services, appear across multiple topics, indicating considerable semantic overlap. These topics are therefore better interpreted as mixtures of surrounding place contexts than as clearly distinct activity purposes.

Figure \ref{fig:lda_tba_clusters} shows the cluster-level topic distributions. Although LDA-TBA identifies eight clusters, several are dominated by the same topic and differ mainly in its relative weight. In particular, Topic~7 is prominent across Clusters~1--4, 6, and~8, while the largest group, Cluster~7, has a comparatively diffuse topic distribution. This suggests that the solution partly fragments similar semantic profiles rather than recovering clearly differentiated activity patterns. Moreover, because LDA-TBA aggregates time allocation at the individual level, it does not retain the timing or sequential organisation of activities. It can distinguish broad differences in visited place contexts, but provides limited evidence about how activities are
temporally organised into distinct daily routines.

\subsubsection{SE-AC}

\begin{figure}
    \centering
    \includegraphics[width=\linewidth]{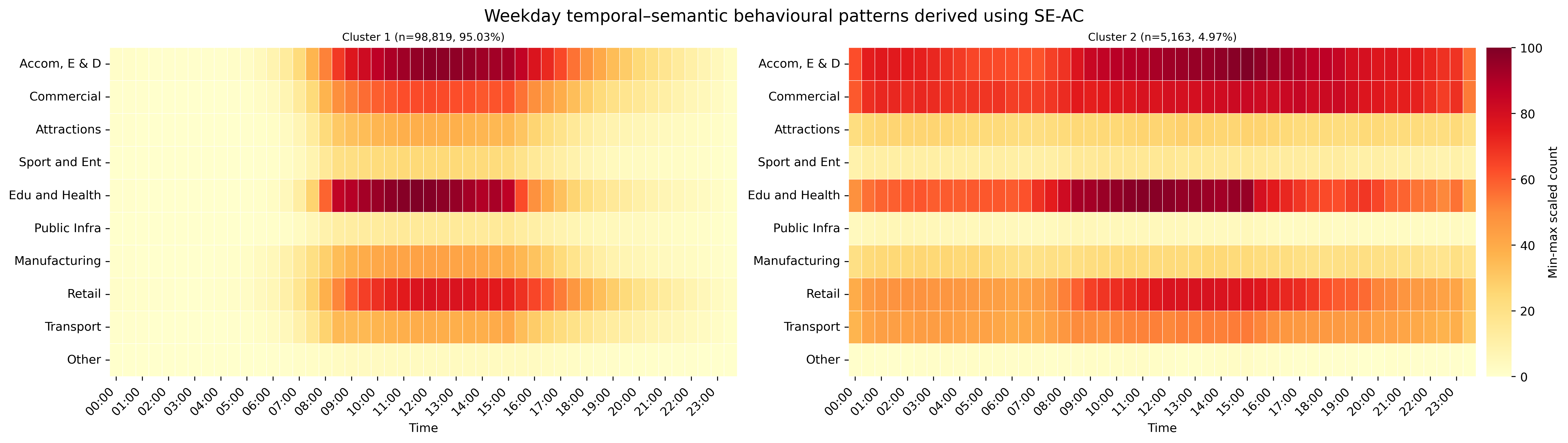}
    \caption{Weekday temporal-semantic activity patterns of the clusters derived using SE-AC. 
    }
    \label{fig:se_ac_clusters}
\end{figure}

Figure~\ref{fig:se_ac_clusters} presents the weekday temporal--semantic patterns obtained using SE-AC. The clustering solution is highly imbalanced: Cluster~1 contains 95.03\% of clustered individuals, while Cluster~2 accounts for only 4.97\%. Cluster~1 exhibits a broad daytime pattern, with relatively strong activity in education and health, accommodation, commercial, and retail contexts. Cluster~2 shows more sustained and widely distributed activity across both hours and urban functions. The distinction between the two groups therefore appears to be driven mainly by activity intensity and breadth rather than by fundamentally different temporal--semantic structures. This limited differentiation is consistent with the SE-AC representation, which aggregates activity-type embeddings using Smooth Inverse Frequency and consequently does not retain the timing or sequential organisation of activities within each chain.

\subsubsection{SE-MMT}

\begin{figure}
    \centering
    \includegraphics[width=\linewidth]{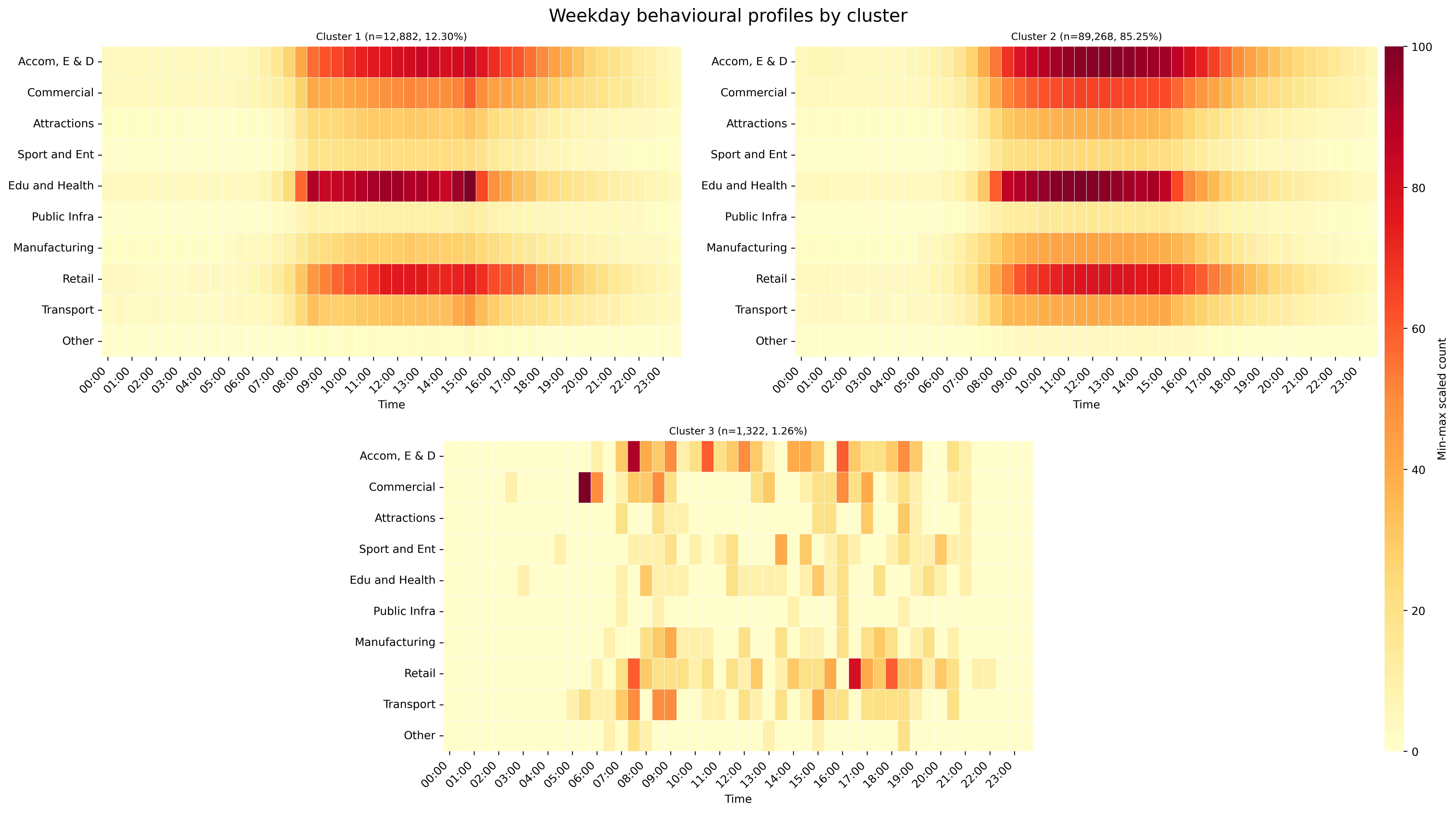}
    \caption{Weekday temporal-semantic activity patterns of the clusters derived using SE-MMT. 
    }
    \label{fig:semmt_cluster_heatmap}
\end{figure}

Figure~\ref{fig:semmt_cluster_heatmap} shows the weekday temporal--semantic patterns derived using SE-MMT. Cluster~2 dominates the solution, accounting for 86.27\% of clustered individuals, compared with 12.45\% for Cluster~1 and 1.28\% for Cluster~3. Clusters~1 and~2 display broadly similar daytime activity structures, particularly across education and health, accommodation, commercial, and retail functions, with their differences expressed mainly through relative intensity and temporal extent. Cluster~3 exhibits a sparse and irregular profile and represents only a very small share of the population. Although SE-MMT uses a pre-trained BERT model, its semantic sentence records activity types and their proportions rather than their original timing and order. The resulting representation therefore captures differences in activity composition but provides limited separation of distinct daily routines.

\subsubsection{Overall comparison}

The baseline methods exhibit different but substantial limitations. LDA-TBA produces relatively fragmented clusters from semantically overlapping and often weakly interpretable topics; moreover, its time-budget aggregation removes the sequential and within-day temporal structure needed to derive behaviourally interpretable temporal--semantic profiles. SE-AC compresses most individuals into a single broad cluster because its SIF aggregation does not retain activity timing or order, while SE-MMT provides only slightly greater differentiation and remains dominated by one general profile, as its textual representation primarily captures semantic composition rather than detailed spatiotemporal information. Consequently, the baseline results offer limited practical value for identifying distinct population activity patterns. In comparison, ACE yields a more differentiated and interpretable set of clusters by jointly preserving the spatial, temporal, and sequential context of daily activity chains.

\section{Discussion}

Despite decades of research and increasing adoption in planning and policy, analysing human mobility from mobile phone data remains inherently challenging. Human mobility exhibits regularity and stochastic variation simultaneously, while the interactions between spatial location, temporal context, activity duration, and urban function add further complexity \citep{song2010limits,hong2023context,wang2023would}. Mobile phone data, meanwhile, provide only indirect and incomplete observations of these behavioural processes. This study responds to these challenges by conceptualising activity pattern mining as a three-stage analytical process: learning representations of individual mobility behaviour, clustering the resulting representations to identify prominent population groups, and analysing and interpreting their temporal, semantic, and demographic characteristics. This formulation positions representation learning as the critical link between raw mobility observations and meaningful population-level knowledge.

Methodologically, ACE introduces two core innovations. The first is the use of pre-trained urban embeddings to represent the geographic context of observed activities. Learned from large-scale multimodal geographic data, such as satellite imagery and points of interest, these embeddings encode rich spatial-semantic information about urban environments \citep{liu2026cityrep}. They also reflect the inherently mixed-use nature of cities, where individual locations may simultaneously support multiple urban functions \citep{yue2017measurements}. In this respect, urban embeddings play a role similar to probabilistic place representations derived from topic models such as LDA: both quantify locations using continuous vectors and avoid assigning each location to a single deterministic activity category. However, pre-trained urban embeddings provide more compact, information-rich, and expressive representations by integrating multiple geographic data sources and capturing nonlinear characteristics of urban environments. Their successful application in quantitative human mobility modelling provides further evidence for their transferability to activity pattern analysis \citep{wang2025into}.

The second innovation is the use of self-supervised learning to address the absence of activity-purpose and behavioural-class labels. Under this general principle, ACE incorporates two complementary learning objectives. Masked activity modelling encourages the model to recover masked activity information from the surrounding sequence, thereby learning contextual dependencies within daily activity chains. The contrastive objective aligns representations of chains associated with the same individual and day type while distinguishing them from other behavioural observations. Together, these objectives exploit regularities inherent in the mobility data and enable ACE to jointly encode urban context, temporal organisation, activity duration, and sequential relationships without relying on externally assigned labels.

Nevertheless, several fundamental limitations of mobile phone data remain. First, all such datasets contain sampling and observation biases to varying degrees. The observed users may not represent the wider population evenly, while data availability may differ systematically across users, locations, and times. The patterns identified in this study may therefore inherit biases from the underlying dataset and should not be interpreted as fully representative of the entire London population. Such bias is not unique to the present study, but is an unavoidable concern in mobile phone-based mobility analysis \citep{wang2022zooming}.

Second, mobile phone datasets commonly lack reliable ground-truth information about activity purposes, travel modes, demographic attributes, and behavioural group membership. This label scarcity limits the use of supervised learning and makes direct quantitative validation difficult. In this study, the demographic characteristics associated with inferred home locations provide contextual evidence rather than verified individual-level attributes. They should therefore be interpreted as demographic contexts associated with the identified clusters, rather than as known characteristics of individual mobile phone users.

The absence of ground truth also explains why the evaluation is primarily qualitative and comparative. Unlike predictive tasks such as next location prediction, unsupervised activity pattern discovery has no known test target, true cluster assignment, or correct number of behavioural groups. Evaluation must therefore consider the coherence, distinctiveness, and interpretability of inferred patterns, while the credibility of the proposed method is established through systematic comparison with patterns generated using alternative methods \citep{shi2023capturing}. 


Third, the observations are temporally sparse and incomplete. Locations may not be recorded continuously because of signalling conditions, application usage, or users disabling location services. An observed daily activity chain is therefore not equivalent to a complete activity diary, as some stays, journeys, and transitions may be missing. Sparsity also occurs at the individual level, because many users appear inconsistently across the study period and continuous multi-day records are uncommon.

These two levels of sparsity directly shaped the methodological framework. Modelling multi-day activity chains would require a level of temporal continuity that the data rarely provide. We therefore treat the daily activity chain as the primary modelling unit, aggregate the resulting representations into individual-level profiles, and subsequently identify prominent patterns at the population level. This strategy reduces reliance on continuously observed multi-day trajectories and allows information from multiple partial daily records to contribute to a more stable individual representation. Nevertheless, incomplete or spatially truncated daily trajectories may still affect the identified patterns, potentially contributing to the seemingly anomalous profiles of Clusters 3 and 6.

Beyond these data-specific limitations, it is also important to distinguish the general methodological principles underlying ACE from their particular implementation in this case study. Activity-pattern mining necessarily involves a sequence of modelling and analytical stages, including mobility preprocessing, activity-chain construction, representation learning, user-level aggregation, dimensionality reduction, clustering, and interpretation. The specific choices within these stages may reasonably vary across datasets according to their spatial coverage, observation density, sparsity, and sampling characteristics. ACE should therefore be understood not as a rigidly fixed neural architecture, but as an implementation of broader methodological principles: incorporating rich pre-trained representations of the urban environment into activity-chain modelling and learning behavioural representations through self-supervised objectives without requiring deterministic activity-purpose labels. In this study, these principles are instantiated using AETHER embeddings, masked activity modelling, and identity-guided contrastive learning, drawing on established practices in representation learning \citep{devlin2019bert,radford2021clip,baevski2020wav2vec}. Other pre-trained urban embedding models could nevertheless be substituted, while alternative masking, reconstruction, contrastive, or sequence-encoding strategies may be appropriate for mobility datasets with different observation characteristics and analytical objectives. The open-source implementation consequently exposes the principal architectural and training choices through configuration files, providing a transparent basis for researchers to reproduce, evaluate, adapt, and extend the framework across datasets with different observation characteristics.

\section{Conclusion}

Understanding how people organise their daily activities is important for travel behaviour research and urban planning. Although mobile phone location data offer large-scale and longitudinal observations of mobility, their lack of activity-purpose labels and incomplete observation of individual behaviour make activity pattern analysis particularly challenging.

This study addressed this challenge by introducing the Activity Chain Encoder (ACE), a self-supervised representation learning method for unlabelled daily activity chains. ACE combines pre-trained urban embeddings with temporal and duration information and uses a Transformer architecture to model contextual dependencies among activities. It thereby learns representations that preserve spatial-semantic, temporal, duration, and sequential information without relying on predefined activity-purpose labels. Within the broader representation–clustering–interpretation workflow, these daily representations are aggregated into user-level activity profiles, clustered, and interpreted through temporal-semantic and demographic evidence.

Applied to mobile phone data from London, the proposed pipeline identified six distinct weekday activity-pattern groups with interpretable temporal--semantic characteristics and differentiated demographic contexts. Comparisons with LDA-TBA, SE-AC, and SE-MMT showed that alternative representations produced either fragmented clusters based on weakly interpretable topics or overly broad groups that obscured meaningful behavioural variation. These findings demonstrate that representation quality is central to unsupervised activity-pattern discovery and support the development of empirically grounded activity-pattern personas.

Overall, this study provides a conceptual and methodological foundation for deriving behaviourally meaningful population patterns from unlabelled mobile phone location data and an empirical account of heterogeneous activity patterns in London. The open-source implementation supports reproducibility and further adaptation. Future research could develop architectures and learning objectives tailored to sparse and irregular observations, explicitly model uncertainty arising from incomplete activity chains, and evaluate the robustness and generalisability of the approach across comparable datasets and urban context.




\section*{Acknowledgements}

This work was partially supported by the project ``Understand the Impact of COVID-19 \& NPIs on Mobility and Places in Singapore'', funded by The Alan Turing Institute and DSO National Laboratories of Singapore. The first author was jointly funded by UCL Dean's Prize and the China Scholarship Council (No. 202106270039).


\appendix
\section{Appendix}
\label{app:main_sec}

\setcounter{figure}{0}
\renewcommand{\thefigure}{A.\arabic{figure}}
\renewcommand{\theHfigure}{appendix.A.\arabic{figure}}

\setcounter{table}{0}
\renewcommand{\thetable}{A.\arabic{table}}
\renewcommand{\theHtable}{appendix.A.\arabic{table}}

\subsection{Geographical distribution of probed urban functions}
\label{app:sec_geo_dist_probing}

\begin{figure}[!ht]
    \centering
    \includegraphics[width=0.8\linewidth]{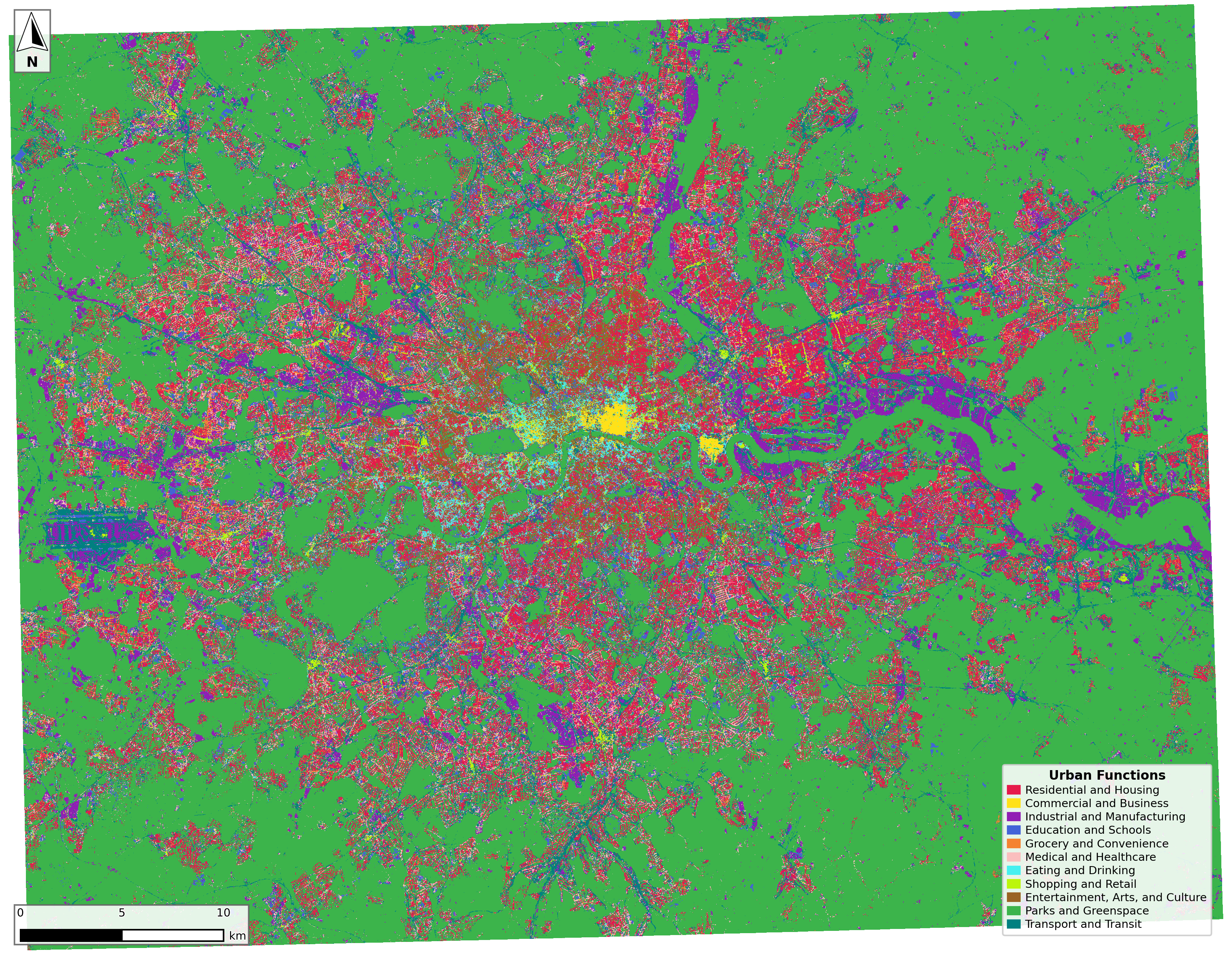}
    \caption{Geographical distribution of the highest-scoring urban functional category derived through semantic probing. Each raster cell is assigned the category with the highest probing score for visualisation only; the full semantic distribution is retained in all subsequent analyses.}
    \label{app_fig:urban_functions}
\end{figure}

The geographical distribution (Figure \ref{app_fig:urban_functions}) shows broad correspondence with the known functional structure of London. Residential and Housing is prevalent across much of the built-up area, while Commercial and Business is strongly concentrated in the City of London and Canary Wharf, two major business districts. Industrial and Manufacturing is more prominent in established industrial areas, particularly in eastern London, whereas Eating and Drinking and Entertainment, Arts and Culture are concentrated around central London, including Soho and the West End. Parks and Greenspace dominates large non-built-up areas, including major parks, open spaces, and water bodies (e.g., the River Thames), consistent with the broad definition of this category. Overall, the strong spatial correspondence between the probed functions and the known urban environment provides a sound basis for the subsequent activity-pattern analysis.

\subsection{Detailed demographic attributes}
\label{app:sec_demo_attr}

\begingroup
\small
\setlength{\tabcolsep}{5pt}
\renewcommand{\arraystretch}{1.12}

\begin{longtable}{
    @{}
    >{\centering\arraybackslash}p{0.08\linewidth}
    >{\raggedright\arraybackslash}p{0.87\linewidth}
    @{}
}
\caption{Classification of demographic variables used in the cluster demographic analysis.}
\label{tab:demographic_categories}
\\

\toprule
\textbf{Code} & \textbf{Category definition} \\
\midrule
\endfirsthead

\multicolumn{2}{l}
{\tablename\ \thetable{} continued from the previous page} \\
\toprule
\textbf{Code} & \textbf{Category definition} \\
\midrule
\endhead

\midrule
\multicolumn{2}{r}{\footnotesize Continued on the next page} \\
\endfoot

\bottomrule
\endlastfoot

\multicolumn{2}{@{}l}{\textbf{Age group}} \\
\addlinespace[0.25em]

1 & Aged 15 years and under \\
2 & Aged 16 to 24 years \\
3 & Aged 25 to 34 years \\
4 & Aged 35 to 49 years \\
5 & Aged 50 to 64 years \\
6 & Aged 65 years and over \\

\addlinespace[0.6em]
\multicolumn{2}{@{}l}{\textbf{National Statistics Socio-economic Classification (NS-SEC)}} \\
\addlinespace[0.25em]

1 & L1, L2 and L3: Higher managerial, administrative and professional occupations \\
2 & L4, L5 and L6: Lower managerial, administrative and professional occupations \\
3 & L7: Intermediate occupations \\
4 & L8 and L9: Small employers and own-account workers \\
5 & L10 and L11: Lower supervisory and technical occupations \\
6 & L12: Semi-routine occupations \\
7 & L13: Routine occupations \\
8 & L14.1 and L14.2: Never worked and long-term unemployed \\
9 & L15: Full-time students \\

\addlinespace[0.6em]
\multicolumn{2}{@{}l}{\textbf{Occupation}} \\
\addlinespace[0.25em]

1 & Managers, directors and senior officials \\
2 & Professional occupations \\
3 & Associate professional and technical occupations \\
4 & Administrative and secretarial occupations \\
5 & Skilled trades occupations \\
6 & Caring, leisure and other service occupations \\
7 & Sales and customer service occupations \\
8 & Process, plant and machine operatives \\
9 & Elementary occupations \\

\addlinespace[0.6em]
\multicolumn{2}{@{}l}{\textbf{Education level}} \\
\addlinespace[0.25em]

0 & No qualifications \\

1 & Level 1 and entry-level qualifications: one to four GCSEs at grades A* to C; 
any GCSEs at other grades; O levels or CSEs at any grade; one AS level; 
NVQ Level 1; Foundation GNVQ; or Basic or Essential Skills qualifications \\

2 & Level 2 qualifications: five or more GCSEs at grades A* to C or 9 to 4; 
O-level passes; CSE grade 1; School Certificate; one A level; two to three 
AS levels; VCEs; Intermediate or Higher Diploma; Welsh Baccalaureate 
Intermediate Diploma; NVQ Level 2; Intermediate GNVQ; City and Guilds Craft; 
BTEC First or General Diploma; or RSA Diploma \\

3 & Level 3 qualifications: two or more A levels or VCEs; four or more AS levels; 
Higher School Certificate; Progression or Advanced Diploma; Welsh Baccalaureate 
Advanced Diploma; NVQ Level 3; Advanced GNVQ; City and Guilds Advanced Craft; 
ONC; OND; BTEC National; or RSA Advanced Diploma \\

4 & Level 4 qualifications or above: bachelor's degree, such as BA or BSc; 
higher degree, such as MA or PhD; PGCE; NVQ Levels 4--5; HNC; HND; 
RSA Higher Diploma; BTEC Higher-level qualifications; or professional 
qualifications, such as teaching, nursing or accountancy qualifications \\

5 & Other qualifications: apprenticeships; vocational or work-related 
qualifications; other qualifications obtained in England or Wales; or 
qualifications obtained outside England or Wales for which the equivalent 
level is not stated or is unknown \\

\addlinespace[0.6em]
\multicolumn{2}{@{}l}{\textbf{Commute distance}} \\
\addlinespace[0.25em]

1 & Less than 5 km \\
2 & 5 km to less than 10 km \\
3 & 10 km to less than 30 km \\
4 & 30 km to less than 60 km \\
5 & 60 km and over \\
6 & Works mainly from home \\
7 & Works mainly at an offshore installation, in no fixed place, or outside the UK \\

\end{longtable}

\noindent
\footnotesize
\textit{Note:} Category codes are specific to each demographic dimension and
therefore restart within each section.
\endgroup

\subsection{Topics identified through LDA-TBA}


\label{app:lda_topics}

\small
\setlength{\tabcolsep}{6pt}
\renewcommand{\arraystretch}{1.15}

\begin{longtable}{
    @{}
    >{\centering\arraybackslash}p{0.07\textwidth}
    >{\RaggedRight\arraybackslash}p{0.88\textwidth}
    @{}
}
\caption{LDA topics and their highest-weighted POI classes. Values in
parentheses indicate the topic word probabilities.}
\label{tab:lda_topics}
\\

\toprule
\textbf{Topic} & \textbf{Highest-weighted POI classes} \\
\midrule
\endfirsthead

\multicolumn{2}{l}{\small\textit{Table \thetable\ continued from the previous page}}\\
\toprule
\textbf{Topic} & \textbf{Highest-weighted POI classes} \\
\midrule
\endhead

\midrule
\multicolumn{2}{r}{\small\textit{Continued on the next page}}\\
\endfoot

\bottomrule
\endlastfoot

1 &
Bus Stops (0.090);
Hair and Beauty Services (0.064);
Convenience Stores and Independent Supermarkets (0.054);
Electrical Features (0.049);
Fast Food and Takeaway Outlets (0.046);
Restaurants (0.040);
Cash Machines (0.029);
Paypoint Locations (0.027);
Cafes, Snack Bars and Tea Rooms (0.025);
Letter Boxes (0.023)
\\

2 &
Bus Stops (0.053);
Cash Machines (0.046);
Clothing (0.039);
Restaurants (0.033);
Fast Food and Takeaway Outlets (0.027);
Accountants and Auditors (0.026);
Computer Systems Services (0.023);
Jewellery and Fashion Accessories (0.022);
Employment Agencies (0.019);
Travel Agencies (0.017)
\\

3 &
Electrical Features (0.090);
Bus Stops (0.059);
Vehicle Repair, Testing and Servicing (0.058);
DIY and Home Improvement (0.046);
Cash Machines (0.029);
Convenience Stores and Independent Supermarkets (0.028);
Vehicle Cleaning Services (0.025);
Petrol and Fuel Stations (0.024);
Unspecified Works or Factories (0.022);
Cafes, Snack Bars and Tea Rooms (0.020)
\\

4 &
Employment Agencies (0.043);
Financial Advice Services (0.042);
Computer Systems Services (0.038);
Solicitors, Advocates and Notaries Public (0.037);
Unspecified and Other Attractions (0.034);
Business Related Consultants (0.033);
Pubs, Bars and Inns (0.025);
Restaurants (0.025);
Cafes, Snack Bars and Tea Rooms (0.025);
Insurers and Support Activities (0.021)
\\

5 &
Cafes, Snack Bars and Tea Rooms (0.068);
Restaurants (0.058);
Fast Food and Takeaway Outlets (0.043);
Cash Machines (0.041);
Bus Stops (0.039);
Hair and Beauty Services (0.039);
Pubs, Bars and Inns (0.037);
Wi-Fi Hotspots (0.029);
Convenience Stores and Independent Supermarkets (0.026);
Property Sales (0.026)
\\

6 &
Fast Food and Takeaway Outlets (0.103);
Hair and Beauty Services (0.070);
Restaurants (0.056);
Convenience Stores and Independent Supermarkets (0.052);
Cafes, Snack Bars and Tea Rooms (0.040);
Bus Stops (0.030);
Cash Machines (0.028);
Pubs, Bars and Inns (0.022);
Property Sales (0.018);
Electrical Features (0.017)
\\

7 &
Electrical Features (0.142);
Bus Stops (0.112);
Letter Boxes (0.052);
Playgrounds (0.034);
Places of Worship (0.028);
Nursery Schools and Pre- and After-School Care (0.028);
Footbridges (0.026);
Building Contractors (0.021);
First, Primary and Infant Schools (0.020);
Bridges (0.018)
\\

8 &
Hair and Beauty Services (0.117);
Restaurants (0.076);
Cafes, Snack Bars and Tea Rooms (0.056);
Property Sales (0.046);
Convenience Stores and Independent Supermarkets (0.024);
Pubs, Bars and Inns (0.021);
Bus Stops (0.021);
Cleaning Services (0.021);
Charity Shops (0.020);
Cash Machines (0.020)
\\

9 &
Clothing (0.145);
Jewellery and Fashion Accessories (0.089);
Restaurants (0.076);
Cafes, Snack Bars and Tea Rooms (0.031);
Footwear (0.023);
Hair and Beauty Services (0.021);
Pubs, Bars and Inns (0.021);
Fast Food and Takeaway Outlets (0.020);
Financial Advice Services (0.020);
Design Services (0.018)
\\

10 &
Hair and Beauty Services (0.103);
Convenience Stores and Independent Supermarkets (0.064);
Fast Food and Takeaway Outlets (0.054);
Bus Stops (0.046);
Grocers, Farm Shops and Pick Your Own (0.039);
Cash Machines (0.038);
Property Sales (0.038);
Restaurants (0.035);
Electrical Features (0.031);
Cafes, Snack Bars and Tea Rooms (0.028)
\\

11 &
Restaurants (0.106);
Hair and Beauty Services (0.048);
Cafes, Snack Bars and Tea Rooms (0.035);
Property Sales (0.034);
Pubs, Bars and Inns (0.034);
Wi-Fi Hotspots (0.028);
Financial Advice Services (0.027);
Convenience Stores and Independent Supermarkets (0.025);
Fast Food and Takeaway Outlets (0.025);
Bus Stops (0.023)
\\

12 &
Clothing (0.089);
Hair and Beauty Services (0.055);
Fast Food and Takeaway Outlets (0.049);
Wi-Fi Hotspots (0.048);
Cafes, Snack Bars and Tea Rooms (0.047);
Cash Machines (0.045);
Restaurants (0.040);
Jewellery and Fashion Accessories (0.030);
Bus Stops (0.026);
Banks and Building Societies (0.026)
\\

13 &
Bus Stops (0.076);
Headquarters, Administration and Central Offices (0.074);
Clinics and Health Centres (0.060);
Footbridges (0.040);
Cafes, Snack Bars and Tea Rooms (0.037);
Cash Machines (0.033);
Convenience Stores and Independent Supermarkets (0.030);
Hospitals (0.024);
Electrical Features (0.023);
Unspecified and Other Attractions (0.022)
\\

14 &
Bus Stops (0.099);
Electrical Features (0.059);
Unspecified and Other Attractions (0.037);
Moorings and Unloading Facilities (0.035);
Hail-and-Ride Zones (0.034);
Letter Boxes (0.034);
Historic and Ceremonial Structures (0.031);
Design Services (0.022);
Pubs, Bars and Inns (0.019);
Nursing and Residential Care Homes (0.017)
\\

\end{longtable}

\normalsize

\printcredits

\bibliographystyle{cas-model2-names}

\bibliography{xlw_refs}


\end{document}